\documentclass{article}

\usepackage{microtype}
\usepackage{graphicx}
\usepackage{subfigure}
\usepackage{defs_ahn}
\usepackage{hyperref}
\usepackage{mathtools}
\usepackage{amsmath, amsthm, amssymb}
\usepackage{url}
\usepackage{lipsum}
\usepackage{caption}
\usepackage{graphicx}
\usepackage{booktabs}
\usepackage{xargs}
\usepackage{xcolor}
\usepackage{bbm}
\usepackage{enumitem}
\usepackage[title]{appendix}
\usepackage[export]{adjustbox}
\usepackage{paralist}
\usepackage{booktabs} 

\usepackage{hyperref}

\usepackage{booktabs}
\usepackage{placeins}

\usepackage[font=small,labelfont=bf]{caption}
\usepackage[colorinlistoftodos,textsize=tiny]{todonotes}
\usepackage{xargs}  
\usepackage{xcolor}
\newcommandx{\sjn}[2][1=]{\todo[linecolor=red,backgroundcolor=red!5,bordercolor=red,#1]{#2}}
\newcommandx{\gautam}[2][1=]{\todo[linecolor=blue,backgroundcolor=blue!5,bordercolor=blue,#1]{#2}}
\newcommandx{\skand}[2][1=]{\todo[linecolor=green,backgroundcolor=green!5,bordercolor=green,#1]{#2}}

\usepackage[accepted]{icml2021_arxiv}

\icmltitlerunning{Structured World Belief}

\begin{document}

\twocolumn[
\icmltitle{Structured World Belief for Reinforcement Learning in POMDP}

\icmlsetsymbol{equal}{*}

\begin{icmlauthorlist}
\icmlauthor{Gautam Singh}{rutgers}
\icmlauthor{Skand Peri}{rutgers}
\icmlauthor{Junghyun Kim}{rutgers}
\icmlauthor{Hyunseok Kim}{etri}
\icmlauthor{Sungjin Ahn}{rutgers,ruccs}
\end{icmlauthorlist}

\icmlaffiliation{rutgers}{Department of Computer Science, Rutgers University}
\icmlaffiliation{etri}{Electronics and Telecommunications Research Institute}
\icmlaffiliation{ruccs}{Rutgers Center for Cognitive Science}

\icmlcorrespondingauthor{Gautam Singh}{singh.gautam@cs.rutgers.edu}
\icmlcorrespondingauthor{Sungjin Ahn}{sjn.ahn@gmail.com}

\icmlkeywords{Machine Learning, ICML}

\vskip 0.3in
]

\printAffiliationsAndNotice{\icmlEqualContribution} 

\begin{abstract}
Object-centric world models provide structured representation of the scene and can be an important backbone in reinforcement learning and planning. However, existing approaches suffer in partially-observable environments due to the lack of belief states. In this paper, we propose Structured World Belief, a model for learning and inference of object-centric belief states. Inferred by Sequential Monte Carlo (SMC), our belief states provide multiple object-centric scene hypotheses. To synergize the benefits of SMC particles with object representations, we also propose a new object-centric dynamics model that considers the inductive bias of object permanence. This enables tracking of object states even when they are invisible for a long time. To further facilitate object tracking in this regime, we allow our model to attend flexibly to any spatial location in the image which was restricted in previous models. In experiments, we show that object-centric belief provides a more accurate and robust performance for filtering and generation. Furthermore, we show the efficacy of structured world belief in improving the performance of reinforcement learning, planning and supervised reasoning. \href{https://sites.google.com/view/structuredworldbelief}{\texttt{https://sites.google.com/view/\\structuredworldbelief}}
\end{abstract}

\section{Introduction}
There have been remarkable recent advances in object-centric representation  learning~\citep{binding}. Unlike the conventional approaches that provide a single vector to encode the whole scene, the goal of this approach is to learn a set of modular representations, one per object in the scene, via self-supervision. Promising results have been shown for various tasks such as video modeling~\citep{gswm}, planning~\citep{op3}, and systematic generalization to novel scenes~\citep{roots}.

Of particular interest is whether this representation can help Reinforcement Learning (RL) and planning. Although OP3~\citep{op3} and STOVE~\citep{stove} have investigated the potential, the lack of belief states and object permanence in the underlying object-centric dynamics models makes it hard for these models to deal with Partially Observable Markov Decision Processes (POMDP). While DVRL~\citep{dvrl} proposes a neural approach for estimating the belief states using Sequential Monte Carlo (SMC) for RL in POMDP, we hypothesize that the lack of object-centric representation and interaction modeling can make the model suffer in more realistic scenes with multiple stochastic objects interacting under complex partial observability. It is, however, elusive whether these two associated structures, objectness and partial observability, can be integrated in synergy, and if yes then how.  

In this paper, we present Structured World Belief (SWB), a novel object-centric world model that provides object-centric belief for partially observable environments, and methods for RL and planning utilizing the SWB representation. To this end, we first implement the principle of object permanence in our model by disentangling the \textit{presence} of an object file in the belief representation from the \textit{visibility} of the object in the observation. Second, we resolve a hard problem that arises when object permanence is considered under partial observability, i.e., consistently re-attaching to an object which can re-appear in a distant and non-deterministic position after a long occlusion. To solve this, we propose a learning mechanism called \textit{file-slot matching}. Lastly, by integrating the object-centric belief into the auto-encoding SMC objective, our model can gracefully resolve tracking errors while maintaining diverse and likely explanations (or hypotheses) simultaneously. This makes our model work robustly with error-tolerance under the challenging setting of multiple and stochastic objects interacting under complex partial observability.



Our experiment results show that integrating object-centric representation with belief representation can be synergetic, and our proposed model realizes this harmony effectively. We show in various tasks that our model outperforms other models that have either only object-centric representation or only belief representation. We also show that the performance of our model increases as we make the belief estimation more accurate by maintaining more particles. We show these results in video modeling (tracking and future generation), supervised reasoning, RL, and planning.

The contributions of the paper can be summarized as: (1) A new type of representation, SWB, combining object-centric representation with belief states. (2) A new model to learn and infer such representations and world models based on the integration of object permanence, file-slot matching, and sequential Monte Carlo. (3) A framework connecting the SWB representation to RL and planning. (4) Empirical evidence demonstrating the benefits and effectiveness of object-centric belief via SWB.

\section{Background}

\textbf{Object-Centric Representation Learning} 
aims to represent the input image as a set of latent representations, each of which corresponds to an object. 
The object representation can be a structure containing the encoding of the appearance and the position (e.g., by a bounding box)~\citep{air,spair,gnm} as in SPACE~\citep{space} or a single vector representing the segmentation image of an object~\citep{nem,iodine,monet,slotattention}. 
These methods are also extended to temporal settings, which we call object-centric world models (OCWM)~\citep{sqair,silot,scalor,gswm,op3}. In the inference of OCWM, the model needs to learn to track the objects without supervision and the main challenge here is to deal with occlusions and interactions. These existing OCWMs do not provide object permanence and belief states. We discuss this problem in more detail in the next section. 



\textbf{Sequential Monte Carlo} methods \cite{smc} provide an efficient way to sequentially update posterior distributions of a state-space model as the model collects observations. This problem is called \textit{filtering}, or \textit{belief update} in POMDP. In SMC, a posterior distribution, or a belief, is approximated by a set of samples called \textit{particles} and their associated \textit{weights}. Upon the arrival of a new observation, updating the posterior distribution amounts to updating the particles and their weights. This update is done in three steps: (1) sampling a new particle via a proposal distribution, (2) updating the weights of the proposed particle, and (3) resampling the particles based on the updated weights. 

\textbf{Auto-Encoding Sequential Monte Carlo} (AESMC) \citep{aesmc,vsmc} is a method that can learn the model and the proposal distribution jointly by simulating SMC chains. 
The model and the proposal distribution are parameterized as neural networks and thus the proposal distribution can be amortized. The key idea is to use backpropagation via reparameterization to maximize the evidence lower bound of the marginal likelihood, which can easily be computed from the weights of the particles.

\section{Structured World Belief}

We propose to represent the state $\bb_t$ of a POMDP at time $t$ as Structured World Belief (SWB) (hereafter, we simply call it belief). As in SMC, SWB represents a belief $\bb_t := (\bs_t, \bw_t)$ as a set of $K$ particles $\bs_t = (\bs_t^1,\dots,\bs_t^K)$ and its corresponding weights $\bw_t = (\bw_t^1,\dots,\bw_t^K)$. However, unlike previous belief representations, we further structure a particle as a set of \textit{object files}: $\bs_t^k = (\bs_{t,1}^k, \dots,\bs_{t,N}^k)$. An object file is a $D$-dimensional vector and $N$ is the capacity of the number of object files. 

\begin{figure}
    \centering
    \includegraphics[width=0.9\linewidth]{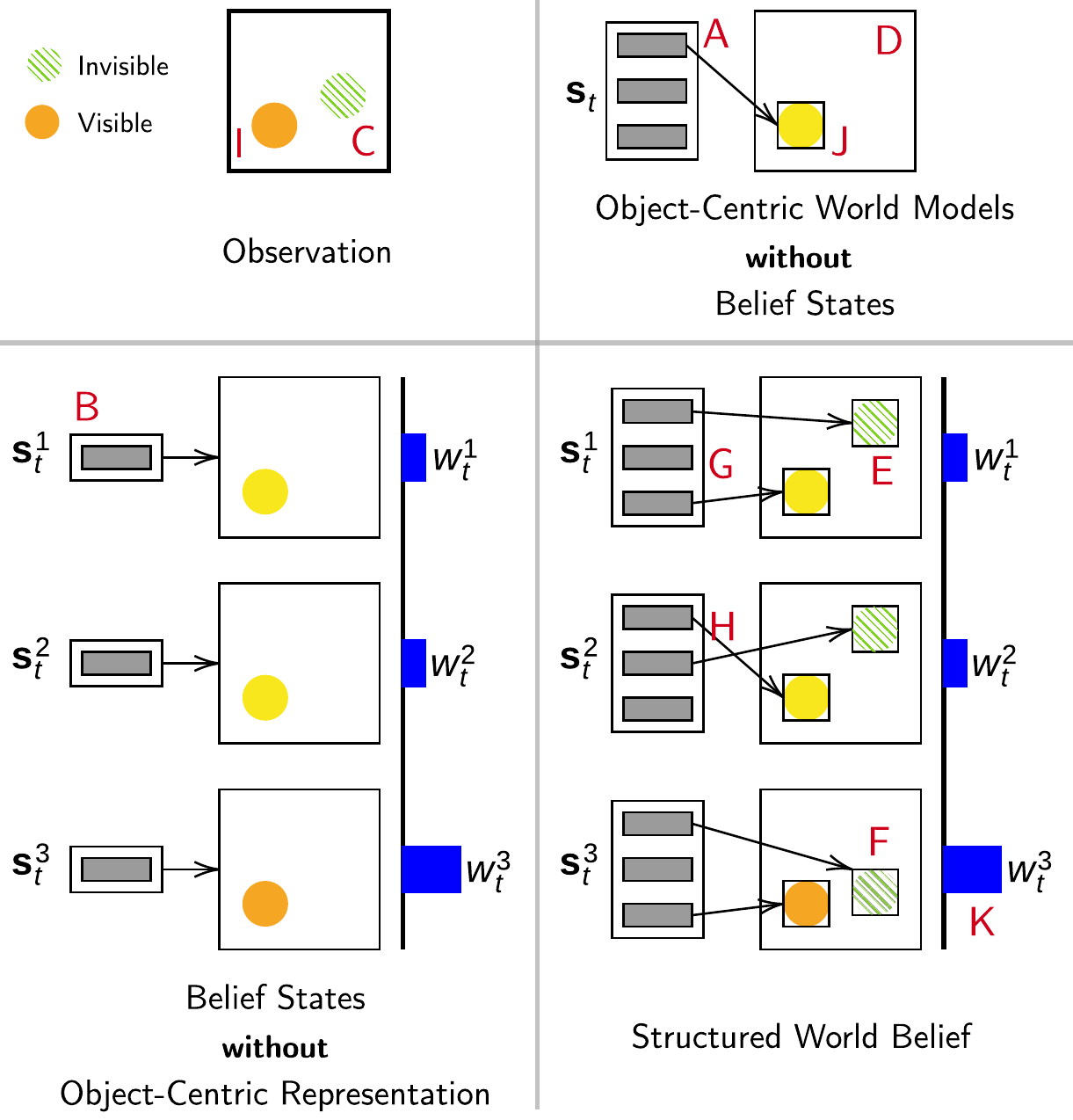}
    \vspace{-2.3mm}
    \small{
    \caption{\textbf{Comparison of Representations.} 
    The grey boxes denote object files and the rectangles (\red{A}, \red{B} and \red{G}) enclosing them denote a scene state or a particle. The figure demonstrates that:
    \textit{i)} object-centric models without belief state maintain single hypothesis (\red{A}) for the scene. 
     \textit{ii)} AESMC maintains multiple hypotheses but the representation lacks object-wise structure (\red{B}). 
     \textit{iii)} Previous object-centric models only maintain files for visible objects (\red{I} and \red{J}) but delete them for invisible objects (\red{C} and \red{D}). \textit{iv)} For invisible objects, SWB has object permanence and maintains multiple hypotheses over its files. The object files \red{E} and \red{F} for the occluded green object provide alternative hypotheses.  
    \textit{v)} In \red{G} and \red{H}, SWB provides belief over different files modeling the same objects. \textit{vi)} If the inference distributions draws a poor sample through chance (for instance, mistaking orange for yellow in \red{J}), previous object-centric models cannot discard this sample. SWB can discard bad samples by assigning high weight to the good samples \red{K}. 
    }
    }
    \label{fig:representation-comparison}
    \vspace{-5mm}
\end{figure}


\textbf{Object Files.} An object file (or simply a file)~\citep{objectfiles,objectfilesandschemata} is a structure containing the following four items:
(1) the ID $i_{t,n}^k$ of the object that the file is bound to, (2) the visibility $\bz_{t,n}^{k, \text{vis}}$ denoting whether the object is \textit{visible} or not, (3) the current state (e.g., appearance and position) of the object $\bz_{t,n}^{k, \text{obj}}$, and (4) an RNN hidden state $\bh_{t,n}^k$ encoding the history of the object. 
All object files are initialized with ID as $\texttt{null}$. This represents that the object files are \textit{inactive}, i.e., not bound to any object. When an object file binds to an object, a positive integer is assigned as the ID meaning that the object file is \textit{activated} and the underlying RNN is initialized. The ID is unique across files $n=1,\dots,N$ within a particle $k$ and thus can be useful in downstream tasks or evaluation.

\textbf{Object Permanence.}~In previous models~\citep{scalor,silot,gswm}, whether a file is maintained or deleted is determined by the visibility of the object. 
When an object is entirely invisible, it does not contribute to the reconstruction error, and thus the file is deleted. That is, these models ignore the inductive bias of object permanence and this makes it difficult to maintain object-centric belief.
To implement object permanence, our model decomposes the presence of an object in the belief and its visibility in the observation. When an object goes out of view, we only set its visibility to zero, but other information (ID, object state, and RNN state) is maintained. This allows for the file to re-bind to the object and set the visibility to 1 when the object becomes visible again. 

\textbf{Object Files as Belief.} When an object becomes invisible, tracking its state becomes highly uncertain because there can be multiple possible states explaining the situation. However, in our model, the active but invisible files corresponding to that object are populated in a set of particles and maintain different states, providing a belief representation of multiple possible explanations. For example, an agent can avoid occluded enemies by predicting all possible trajectories under the occlusion.

\section{Learning and Inference}
\label{sec:swb_learn}

\subsection{Generative Model}
The joint distribution of a sequence of files $\bs_{1:T, 1:N}$ and observations $\bx_{1:T}$ can be described as:
\begin{align}
    p_\ta(\bx_{1:T}, \bs_{1:T}) = \prod_{t=1}^T {p_\ta(\bx_t | \bs_t)} \prod_{n=1}^N {p_\ta(\bs_{t,n} | \bs_{t-1})}\nn
\end{align}
where the file transition $p_\ta(\bs_{t,n} | \bs_{t-1})$ is further decomposed to $p_\ta(\bz_{t,n}| \bz_{t-1}, \bh_{t-1}) p(i_{t,n})$.
Here, $p_\ta(\bz_{t,n}| \bz_{t-1}, \bh_{t-1})$ is the dynamics prior for $\bz_{t,n}^{\text{obj}}$ and $\bz_{t,n}^{\text{vis}}$ and $p(i_{t,n})$ is the prior on ID. 
The RNN is updated by $\bh_{t,n} = \text{RNN}_\ta(\bz_{t,n}, \bh_{t-1,n})$.



\subsection{Inference}

In this section, we describe the inference process of updating $\bb_\tmo$ to $\bb_t$ using input $\bx_t$. At the start of an episode, SWB contains $K\times N$ inactive files and these are updated at every time-step by the input image. This involves three steps: \textit{i)} \textit{file update} from $\bs_{\tmo,n}^k$ to $\bs_{t,n}^k$ and \textit{ii)} \textit{weight update} from $\bw_\tmo$ to $\bw_t$ and \textit{iii)} \textit{particle resampling}. 




\subsubsection{Updating Object Files}

\textbf{Image Encoder and Object Slots.} 
To update the belief, our model takes an input image at each time step. For the image encoder, the general framework of the proposed model requires a network that returns a set of object-centric representations, called \textit{object slots}, $\bu_{t,1:M} = \{\bu_{t,1}, \dots, \bu_{t,M}\}$ for image $\bx_t$. Here, $M$ is the capacity of the number of object slots and we use $m$ to denote the index of slots. To obtain the slots in our experiments, we use an encoder similar to SPACE. However, unlike SPACE that has a separate encoder for the glimpse patch, we directly use $M$ CNN features of the full image which have the highest presence values. These presence values are not necessary if an encoder can directly provide $M$ slots as in \citep{nem, monet, iodine}.
This image encoder is not pretrained but trained jointly end-to-end along with other components.

\begin{algorithm}[t]
\footnotesize
\caption{File-Slot Matching and Glimpse Proposal}
\begin{algorithmic}
\INPUT Previous files $\bs_{t-1}$, Slots $\bu_{t, 0:M}$
\OUTPUT Proposal region $\boldsymbol{o}_{t,n}^{k, \text{proposal}}$ 
\vspace{1.5mm}
\hrule
\vspace{2mm}
\STATE \# \textit{1. Interact each slot with all files}
\STATE $\bee^{k,\text{interact}}_m = \sum_{n=1}^N f^\text{interact}_\phi(\bs_{t-1,n}^k, \bu_{t, m}) \hfill \forall k, m$
\STATE
\STATE \# \textit{2. Compute probabilities to match file to slot}
\STATE $q_{n,m}^{k,\text{match}} = f^\text{match}_\phi(\bs_{t-1,n}^k, \bu_{t, m}, \bee^{k,\text{interact}}_m) \hfill \forall k, n, m$
\STATE 
\STATE \# \textit{3. Obtain matched slot}
\STATE $
m_{t,n}^k
\sim \text{Categorical}(\cdot | q_{n,0:M}^{k,\text{match}})\hfill  \forall k, n$
\STATE
\STATE \# \textit{4. Obtain Glimpse Proposal}
\STATE $\boldsymbol{o}_{t,n}^{k, \text{proposal}} = f_\phi^\text{proposal}(\bs_{t-1,n}^k, \bu_{t, m_{t,n}^k}) \hfill  \forall k, n$
\end{algorithmic}
\label{algo:object_matching}
\end{algorithm}

\begin{figure}
    \centering
    \includegraphics[width=0.9\linewidth]{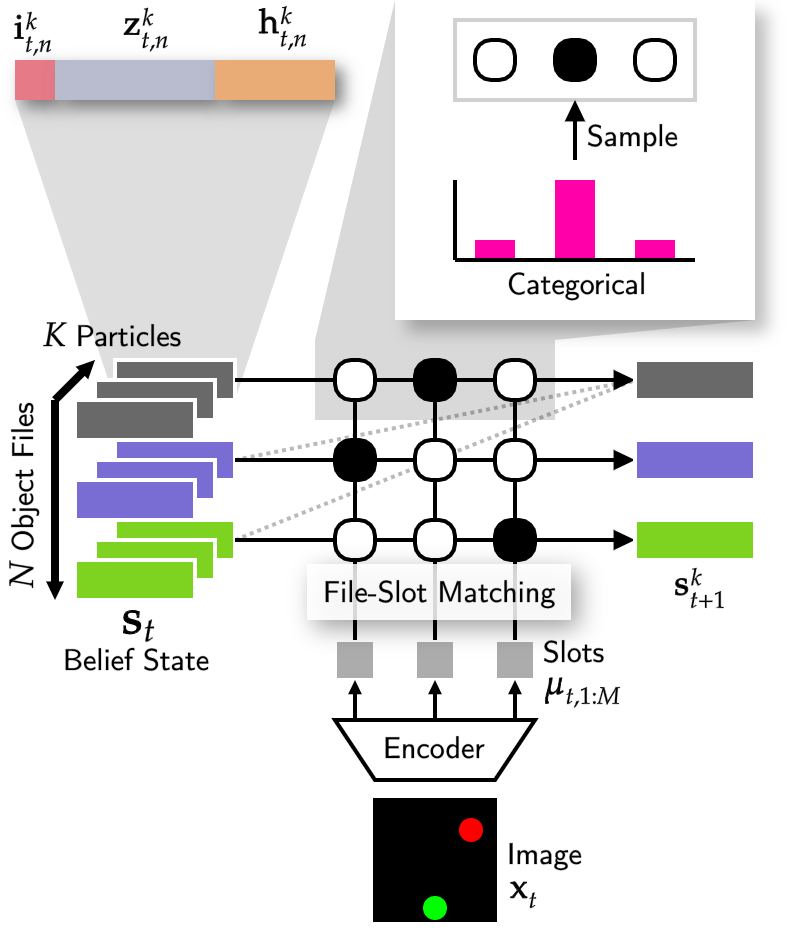}
    \small
    \caption{\textbf{A Simplified View of Inference in SWB.} For each object-file, we sample a matching slot and use it to update the object file. In this update, the object files also interact with each other through a graph neural network (shown in dotted lines).}
    \label{fig:inference_pipeline}
\end{figure}


We introduce a special slot $\bu_{t,0}$, called \textit{null slot} (represented as a zero vector), and append it to the object slots, i.e., $\bu_t = \bu_{t,0} \cup \bu_{t,1:M}$. We shall introduce a process where each object file finds a matching object slot to attend to. The null slot will act as a blank slot to which an object file can match if it is not necessary to attend to any of the object slots $\bu_{t,1:M}$, for example, if the object is active in the file but invisible in the input image.


\textbf{Learning to (Re)-Attach by File-Slot Matching.}
A natural way to update the files is to obtain new information from the relevant area in the input image. In previous works~\citep{silot,scalor,gswm}, this was done by limiting the attention to the neighborhood of the position of the previous object file because an object may not move too far in the next frame. However, this approach easily fails in the more realistic setting of partial observability tackled in this work. For example, when the duration of invisibility of an object is long, the object can reappear in a non-deterministic surprise location far from the previously observed location. Therefore, finding a matching object that can re-appear anywhere in the input image is a key problem.


To address this, we propose a mechanism for learning-to-match the object files and the object slots. Crucially, our model learns to infer the matching implicitly via the AESMC learning objective without any inductive bias. The aim of the matching mechanism is to compute the probability $q_{n,m}^k$ that file $n$ matches with slot $m$. 
To compute this, if each file looks at the candidate slot alone, multiple files can match with the same slot causing false detections or misses during tracking. To prevent this, a file should not only look at the candidate slot but also at other files that are competing to match with this slot. For this, we compute an interaction summary $\bee_m^{k,\text{interact}}$ of interaction between a slot $\bu_{t,m}$ and all the files $\bs_{t-1,1:N}^k$ via a graph neural network. After doing this for all $m$ in parallel, we compute the probability of matching $q_{n,m}^k$ using an MLP that takes as input the interaction summary, the file $n$, and the slot $m$. For each file, we then apply softmax over the slots. This constructs a categorical distribution with $M+1$ output heads. From this distribution, we sample the index $m_{t,n}^k$ of the matching slot for the file $n$. The randomness in this sampling, when combined with SMC inference, helps maintain multiple hypotheses. The match index $m_{t,n}^k=0$ denotes a match to the null slot.

\textbf{State Inference.}
Given the matched slot, it is possible to directly use the encoding of the slot $\bu_{t,m_{t,n}^k}$ provided by the encoder to update file $n$. However, we found it more robust to use the slot $\bu_{t,m_{t,n}^k}$ and the file $\bs_{t-1, n}^k$ to first infer a glimpse area via an MLP network. The image crop from the glimpse area is then encoded as $\bee_{t,n}^k$ and used to infer the object visibility, the object state, and the RNN state of the file $n$: 
\begin{align}
    \bz_{t,n}^{k, \text{vis}} &\sim \text{Bernoulli}_\phi(\bz_{t,n}^{k, \text{vis}} | \bs_{t-1}^{k}, \bee_{t,n}^k),\nn\\
    \bz_{t,n}^{k, \text{obj}}  &\sim q_{\ta,\phi}(\bz_{t,n}^{k, \text{obj}} | \bs_{t-1}^k, \bee_{t,n}^k, \bz_{t,n}^{k, \text{vis}}),\nn\\
\bh_{t,n}^k &= \text{RNN}_\ta(\bz_{t,n}^{k}, \bh_{t-1,n}^k).\nn
\end{align}



\textbf{Imagination.} A crucial advantage of our object-centric representation is that we can control the belief at the individual object-level rather than the full scene level. This is particularly useful when we update the state of an invisible object file while other files are visible. For the invisible object, the input image does not provide direct information for updating the state of the invisible file and thus it is more effective to use the prior distribution for the invisible files instead of letting the posterior re-learn the dynamics of the prior. We call this update based on the prior as the imagination phase. Note that the objects governed by the prior can still interact with other objects, both visible and invisible. To allow this transition between imagination and attentive tracking at the individual object level, we factorize $q_{\ta,\phi}(\bz_{t,n}^{k, \text{obj}} | \bs_{t-1}^k, \bx_t, \bz_{t,n}^{k, \text{vis}})$ according to the visibility:
\begin{align}q_\phi(\bz_{t,n}^{k, \text{obj}} | \bs_{t-1}^k, \bx_t)^{\bz_{t,n}^{k, \text{vis}}}p_\ta(\bz_{t,n}^{k, \text{obj}} | \bs_{t-1}^k)^{1 - {\bz}_{t,n}^{k, \text{vis}}}\nn.
\end{align}

\textbf{Discovery and Deletion.} The discovery of an object is the state transition from inactive to active which is learned through the file-slot matching. An active file carries over its assigned ID until it becomes inactive. Even if a file is invisible for a long time, we do not explicitly delete an object as long as the file state stays active. Rather, these files are updated via imagination. Alternatively, we can introduce the invisibility counter, and when the file capacity is near full, we can delete the file which has been invisible for the longest time.

\subsubsection{Weight Update and Resampling}
The object-centric decomposition in our model allows the following decomposition of the weight computation:
\begin{align}
    w_t^k \propto w_{t-1}^k\cdot  p_\ta(\bx_t|\bz_{t}^k)\prod_{n=1}^N w_{t,n}^{k, \text{file}} \cdot w_{t,n}^{k, \text{ID}}  \cdot w_{t,n}^{k, \text{match}}\nn
    \label{eq:swb-reweigh}
\end{align}
where
\begin{align}
    w_{t,n}^{k,\text{file}} &= \dfrac{p_\ta(\bz_{t,n}^{k, \text{obj}} | \bs_{t-1}^k) }{q_{\ta,\phi}(\bz_{t,n}^{k, \text{obj}} | \bs_{t-1}^k, \bx_t, \bz_{t,n}^{k, \text{vis}})}\dfrac{p_\ta(\bz_{t,n}^{k, \text{vis}}| \bs_{t-1}^k) }{q_\phi(\bz_{t,n}^{\text{vis},k} | \bs_{t-1}^k, \bx_t)},\nn\\
    w_{t,n}^{k, \text{ID}} &=\dfrac{p(i_{t,n}^k) }{q_\phi(i_{t,n}^k | \bs_{t-1}^k, \bx_t)},\nn\\
    w_{t,n}^{k, \text{match}} &= \dfrac{p(m_{t,n}^k) }{q_\phi(m_{t,n}^k | \bs_{t-1}^k, \bx_t)}\nn.
\end{align}
To allow object files to match to any slot without an inductive bias such as spatial locality in previous works, we apply a uniform prior $p(m_{t,n}^k)$ on the matching index latent. Using these updated weights, we then perform resampling. 

\subsection{Learning Objective}
Our learning objective is the same Evidence Lower Bound (ELBO) used in AESMC which can be computed by:
$\mathcal{L}_{\ta,\phi} = \frac{1}{T}\sum_{t=1}^T \log \sum_{k=1}^K w_t^k.$
To optimize the objective via gradient descent, we backpropagate through the random variables via the reparameterization trick \cite{vae} and for discrete random variables, we use REINFORCE \cite{reinforce}. We found the straight-through estimator to be less stable than REINFORCE. 

\section{Structured World Belief Agent}

To use our structured world belief for the downstream tasks such as RL, planning, and supervised learning, we need to encode the particles and weights into a belief vector. For this, we first encode each object file in the belief using a shared MLP and obtain $N\times K$ such file encodings. Then, we apply mean-pooling over objects to obtain $K$ particle encodings. To each particle encoding, we concatenate the particle weight. These are fed to another shared MLP to obtain $K$ particle encodings, to which we again apply a mean-pooling and apply a final MLP to obtain the final encoding of the structured world belief.

\section{Related Works}\label{sec:related}

\textbf{Neural Belief Representations}: There are several works that use belief states for RL and planning in POMDP \cite{kaelbling1998planning}. DRQN \cite{hausknecht2015deep} and ADRQN \cite{zhu2017improving} use RNN states as input to the Q-network. RNN history as the sufficient statistics of the belief has also been used in  \cite{moreno2018neural} for supervised reasoning tasks, in \cite{guo2018neural, simcore, PBL} for sample-efficient RL, and in \cite{gangwani2020learning} for imitation learning.
PF-RNNs \cite{pfrnn} incorporate belief as hidden state distribution in the RNN.
\cite{tschiatschek2018variational, wang2019dualsmc, piche2018probabilistic} use belief states for policy optimization in planning.
One of the closest works to SWB is that of DVRL \cite{dvrl}. DVRL combines AESMC~\cite{aesmc}-based world model with A2C \cite{mnih2016asynchronous}, an actor-critic policy gradient method. AlignNet \cite{creswell2020alignnet} uses attention matrix in order to align objects in a scene and track them over time. It also incorporates an object slot memory module in order to capture object permanence demonstrated for short-term occlusions. Unlike SWB, it does not  
provide structured belief for the invisible objects and cannot perform long term future generation. R-NEM \cite{van2018relational} uses a spatial mixture model to obtain structured representation of a scene. While R-NEM implicitly handles object permanence, it does not provide an explicit belief over object states.


\textbf{Object-centric approaches in RL and planning}: Object-centric representations have been considered for use in Reinforcement Learning \cite{scholz14, diuk2008object, mohan2011object, goel2018unsupervised, keramati2018strategic, devin2018deep}. Recent works have focused on self-supervised learning objectives to infer object-centric representations and benefit  Reinforcement Learning.
OP3 \cite{op3} uses
a temporal spatial-mixture model to obtain disentangled representations while O2P2 \cite{janner2019reasoning} uses the ground truth object segmentation mask to learn a predictive model for control. COBRA \cite{watters2019cobra} uses MONet \cite{monet} to obtain object-centric representations. Transporter \cite{transporter} uses KeyNet \cite{jakab2018unsupervised} to extract key-points for sample-efficient RL. \citet{davidson2020investigating} also show that object representations are useful for generalization and sample-efficiency in model-free RL. \citet{liang2015state} and \citet{agnew2020relevanceguided} use handcrafted object features for RL. 

\section{Experiments}

The goal of the experiments is to \textit{i)} evaluate the belief state provided by SWB for modeling the true object states \textit{ii)} evaluate the performance of SWB when it is used as a world model for reinforcement learning and planning and \textit{iii)} evaluate the performance of SWB when the inferred belief states are used as inputs for supervised learning tasks.




\textbf{Environments.} In all the environments, the objects can disappear for 25 to 40 time-steps creating the long-term invisibility that we tackle in this work. We experiment using the following environments: \textit{i) 2D Branching Sprites.} This is a 2D environment with moving sprites. The object paths split recursively, with objects randomly taking one branch at every split. Each sprite is associated with a pair of colors and the color switches periodically. We have two versions of this data set: one in which objects can spawn and another in which objects remain fixed. \textit{ii) 2D Branching Sprites Game.} 
We turn the 2D Branching Sprites environment into a game in which the aim of the agent is to select a lane on which an invisible object is present. The reward and penalty depend on object color. This makes it important for agent to have accurate belief state over both position and appearance. \textit{iii) 2D Maze Game.} In this game, the objective of the agent is to reach a goal and avoid the enemies. \textit{iv) 3D Food Chase Game.} In this game, the objective of the agent is to chase and eat the food and avoid the enemies. For more details of each environment see Appendix \ref{ax:envs}.

In these environments, by having color change and randomness in object dynamics, we test SWB in modeling multiple attributes of invisible objects. The dynamics of color change in 2D Branching Sprites also test the ability of the model in having memory of the past colors. Lastly, the objects can have identical appearances which makes it important for the model to utilize both the position and appearance based cues to solve object matching problem.




\begin{figure*}[h!]
    \centering
    \includegraphics[width=\linewidth]{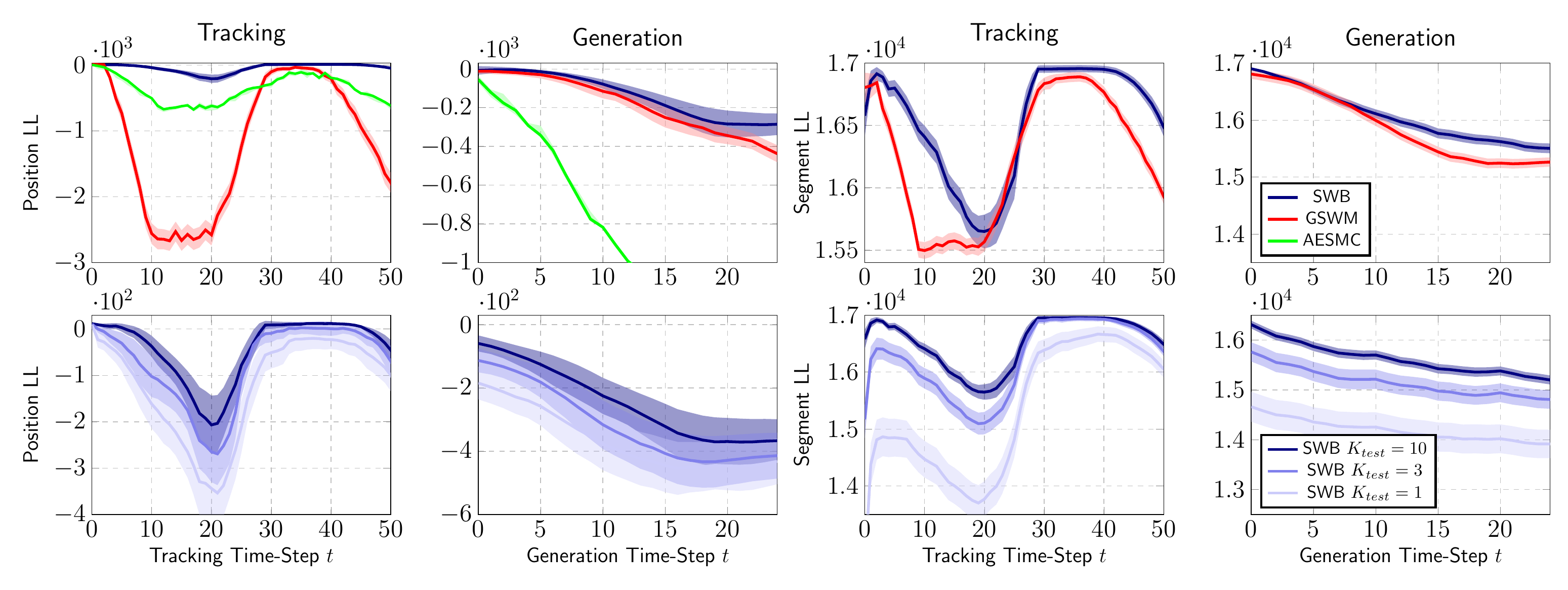}
    \vspace{-8.5mm}
    \caption{Comparison of SWB in terms of position log-likelihood and segment log-likelihood with the baselines: GSWM, AESMC and SWB $K=$1 and 3 on 2D Branching Sprites without new object spawning during the episode.
    }
    \vspace{-1mm}
    \label{fig:all_nlls_nospawn}
\end{figure*}
\begin{figure*}[h!]
    \centering
    \includegraphics[width=1.0\textwidth]{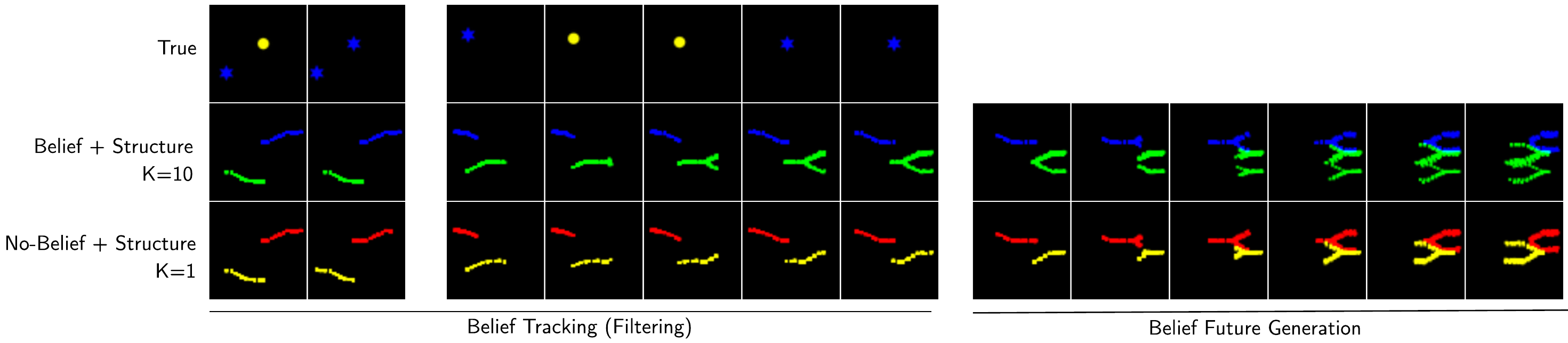}
    \vspace{-6mm}
    \caption{\textbf{Comparison between $K=1$ and $K=10$ for Belief Tracking and Generation in SWB in 2D Branching Sprites.} The top row shows observations and the middle and the bottom rows visualize 8 most recent positions of each active object file. In the beginning, two objects are shown and model assigns object files to each object. Subsequently, one object becomes invisible and other remains visible. We note that the belief using $K=10$ properly represents multi-modal hypotheses for the position of the invisible object in comparison to $K=1$. We also note that imagination is performed only for the invisible object while the state of the visible object is inferred via the matched slot from the input image. In future generation of the belief, we note that starting generation from a belief with $K=10$ particles allows us to generate more complete set of possible futures (see green object file). On the contrary, when the belief is maintained using $K=1$, the set of generated future states do not capture all modes of the possible futures. (see yellow object file)
    }
    \label{fig:qual-2d-branching-sprites-ax}
\end{figure*}

\subsection{World Belief Modeling}
 \label{sec:world_model}
 \subsubsection{Representation Log-Likelihood}
 


We first evaluate the accuracy of the current belief, or in other words, the filtering distribution conditioned on observations seen so far. We expect SWB to provide accurate hypotheses about possible object positions and appearance even when objects are not visible. We note that our task is significantly more challenging than the previous object-centric world models because unlike these works, the objects are fully invisible for much longer time steps.
 

 

\textbf{Model and Baselines.} We evaluate our model SWB with $K=10$ particles which provides both object-centric representation and belief states. We compare against GSWM \cite{gswm} which provides object-centric world representation but does not provides belief states. We also compare against AESMC \cite{aesmc} which provides belief states but not object-centric world representation. Because AESMC does not provide object-wise positions or segments, we train a supervised position estimator to extract position from the AESMC belief representation. Note that the supervised learning signal is only used to train this estimator and is not propagated to AESMC. We also compare against the non-belief version of SWB i.e. SWB with $K=1$ which can be seen as GSWM augmented with object-permanence. This enables it to keep files for invisible objects but the belief state is composed of only one structured state. To also study the effect of gradually increasing the number of particles we compare against SWB with $K=3$. 


\textbf{Metrics.} 
We evaluate our belief using ground truth values of object-wise positions and object segments. 
We use the predicted particles to perform kernel density estimation of the distribution over these attributes and evaluate the log-likelihood of the true states. We compute MOT-A \cite{milan2016mot16} metric components (such as false positives and switches) by counting them for each particle $k$ and taking weighted average using the particle weights $w_t^k$. For more details, see Appendix \ref{ax:metrics}.


\textbf{Tracking.}
We infer and evaluate the current belief on episodes of length 50. For this experiment, we consider the 2D Branching Sprites environment with no object spawning. The objects alternate between being visible and invisible, leading to a periodic variation in number of objects visible at different points of time in tracking. 
In Fig.~\ref{fig:all_nlls_nospawn}, we see that by increasing the number of particles from $K=1$ to 3 and 10, we are able to improve our modeling of density over the object states. Our improvements are more pronounced when the objects are invisible. This is because more particles lead to a diverse belief state that is less likely to miss out possible states of the environment, thereby showing improvement in density estimation. Furthermore, we see that previous object-centric models (evaluated as GSWM) suffer in performance when the objects become invisible due in large part to deletion of their object files. Lastly, we see that in comparison to models providing unstructured belief (evaluated as AESMC with $K=10$), our belief modeling is significantly more accurate. This is enabled by our modeling of structure which allows us to accurately track several possible hypothesis for states of objects.

\begin{figure}[t]
    \vspace{0mm}
    \centering
    \includegraphics[width=0.485\textwidth]{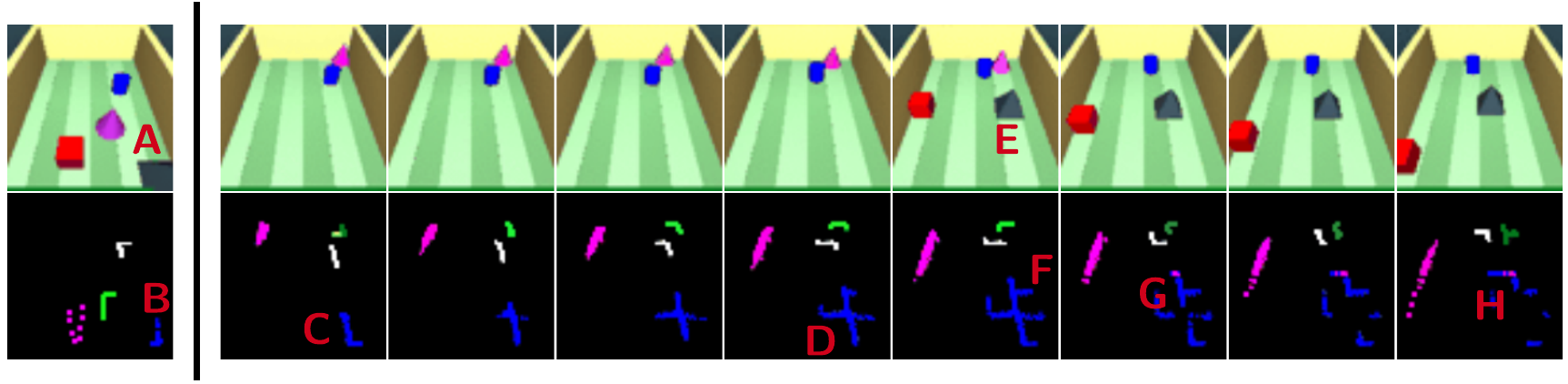}
    \vspace{-8mm}
    \caption{\textbf{Structured Belief in 3D Food Chase}. Top row shows observations and bottom row visualizes 8 most recent position states of each active object file. The color of the file denotes ID. The black pyramid (\red{A}) is tracked by a blue file (\red{B}). The pyramid becomes invisible and file continues to track it (\red{C} through \red{D}). As it remains invisible, the file produces possible hypotheses of its trajectory (\red{D}) which we see as splitting of paths. Hence, SWB can selectively perform imagination for the invisible object and bottom-up inference of visible objects. In \red{E}, The pyramid becomes visible and the file re-attaches to it (\red{F}). We see that the magenta object file also tries to attach to the pyramid (\red{G}). But this hypothesis gets low weight and the blue object file wins the competition (\red{H}).  }
    \label{fig:qual-3d}
\end{figure}

\textbf{Generation.}
In this task, we study how well our model predicts the density over future states of the scene conditioned on the provided observations $\bx_{1:t}$. That is, we seek to evaluate the density $p(\bs_{t+\Delta} | \bx_{1:t})$ where $t + \Delta$ is a future time-step. Due to partial observability, our generation task is more challenging than what previous object-centric world models perform. In previous models, this generation task is equated to predicting future samples from the last inferred scene state $p(\bs_{t+\Delta} | \bs_{t})$. However, when occlusion and object invisibility are present, a single scene state $\bs_{t}$ cannot capture diverse possible states in which the environment can be at time $t$ when the generation is started. 

In Fig.~\ref{fig:all_nlls_nospawn}, we study the effect of number of particles used in SWB for maintaining belief $K=1$ and $K=3$ at the start of generation. During generation, note that we draw the same number of future roll outs to eliminate the effect of number of roll-outs in our kernel density estimation. We observe that for SWB $K=1$ and $K=3$, the generation starts with a poor density estimation which gets worse as the roll-out proceeds. Fig.~\ref{fig:all_nlls_nospawn} compares our future generation against baselines GSWM and AESMC. Since GSWM deletes files when objects are invisible, we made sure that objects were visible in the conditioning period. Because of this, there is no requirement of good belief at the start of the generations. Despite this, we note the generation in GSWM to be worse than that of SWB. We also ensured that same number of generation samples were drawn for all the models compared. Generation in AESMC suffers severe deterioration over time because of the lack of object structure.
\begin{figure}
    \centering
    \includegraphics[width=0.95\linewidth]{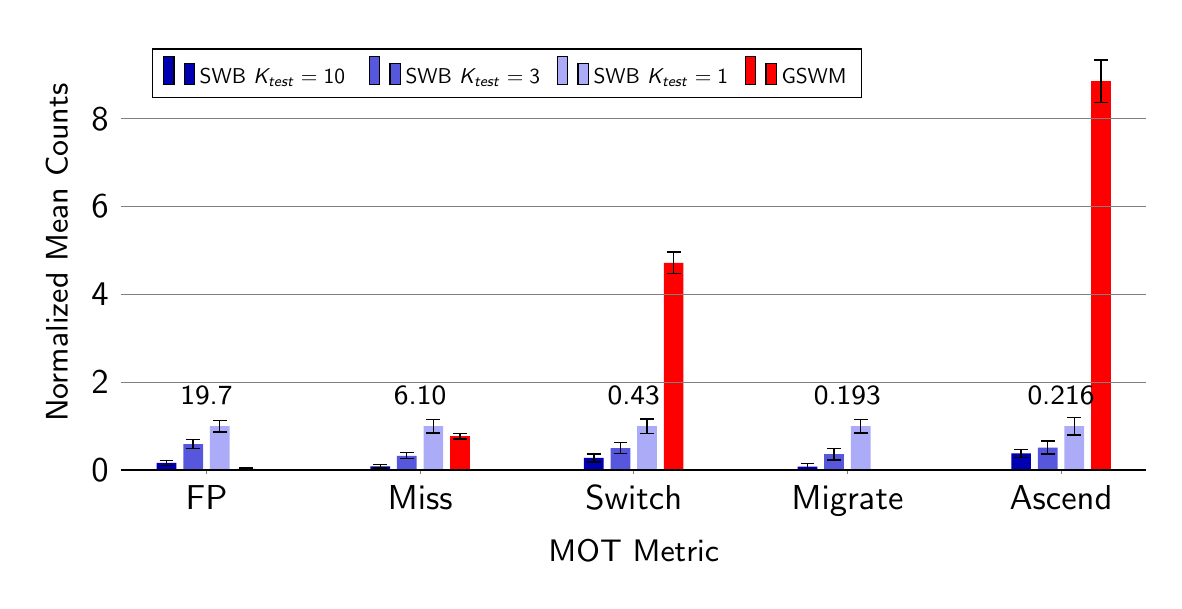}
    \vspace{-6mm}
    \caption{Effect of number of particles used in tracking on the MOT error metrics on Branching Sprites dataset (No-Spawn) with episode length 100. The values are normalized with respect to SWB $K_\text{test}=1$.  Values corresponding to 1 are labelled. The counts are first summed over length of the episode and then averaged over episodes. \textit{FP} refers false positive detections. \textit{Miss} denotes missed detections. \textit{Switch} denotes ID swaps. \textit{Migrate} denotes when an existing file gets assigned to a new object. \textit{Ascend} denotes when a new file is created for an object which already has a file.}
    \label{fig:mota-metrics-analysis}
    \vspace{-4mm}
\end{figure}

    \textbf{MOT.} In Fig.~\ref{fig:mota-metrics-analysis}, we show MOT-A metrics \cite{milan2016mot16} and compare SWB for different number of particles and against GSWM. We note that SWB $K=10$ is the most robust and makes the least errors. This is because by having more particles, even if the inference network makes a few errors in file-slot matching in few particles, those particles receive low weights and get pruned. We see that when we use fewer particles ($K=3$ and 1), the error rate increases because the model cannot recover from the errors. We hypothesize that this is also the reason why GSWM has higher error rate than ours in various metrics. A notable point is the high \textit{ascend} rate in GSWM when a new file gets created for an object which already has a file. This is because GSWM deletes the original file when the object becomes invisible and needs to create a new file when the object re-appears. 

\begin{figure}[t]
    \centering
    \vspace{0mm}
    \includegraphics[width=0.95\linewidth]{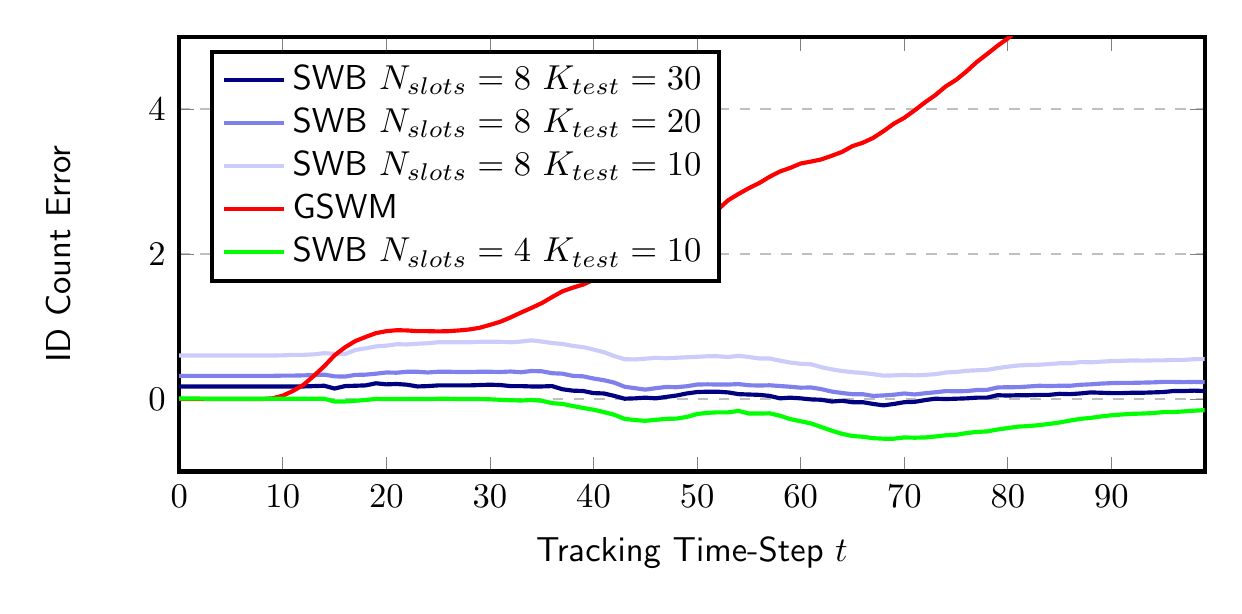}
    \vspace{-5mm}
    \caption{
    Error in number of spawned IDs against true number of objects over the tracking time-steps on 2D Branching Sprites with 4 object spawns during the episode. Negative and positive values denote under-counting and over-counting, respectively.
    }
    \label{fig:slot-generalization}
    \vspace{-2mm}
\end{figure}
\begin{figure*}
    \centering
    \includegraphics[width=1.0\linewidth]{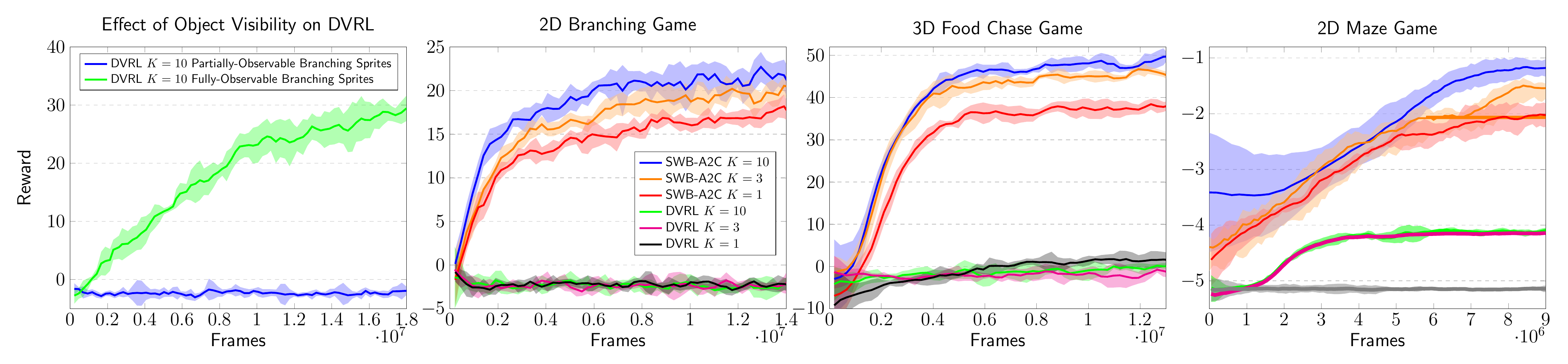}
    \vspace{-9mm}
    \caption{Comparison with DVRL in terms of average reward of A2C when SWB is used as a world model. We report performance for different values of $K$. The SWB world models for 2D Branching Sprites, 3D Food Chase game and 2D Maze Game were pre-trained using frames collected through 200K, 200K and 500K interactions with the environment, respectively, using a random policy.}
    \label{fig:a2c-all}
    \vspace{-2mm}
\end{figure*}
\textbf{Object Spawning, Object Permanence and Robust Generalization.}
In  Fig.~\ref{fig:slot-generalization}, we investigate how our model behaves when objects spawn during the episode, how it generalizes to a larger number of objects in the scene, object files and particles than used in training. For this we trained our model on 2D Branching Sprites with upto 2 objects but we test in a setting in which up to 4 objects can spawn. In  Fig.~\ref{fig:slot-generalization}, we note that when we test our model with $K=10$ and $N=4$ which is the same as training configuration, our model is able to generalize well to 4 objects with small under-counting. The slight under-counting is expected because the model has never seen 4 objects during training. We note that GSWM, due to lack of object permanence, is unable to attach existing files to objects which re-appear after being invisible. Hence, we see an increasing error in the number of object files created in comparison to the true object count. Next, we increase the number of object files in our model from 4 to 8. We see that our model generalizes well and remains robust with slight over-counting. Lastly, we see that we can eliminate this over-counting by increasing the number of particles from $K=10$ to $K=30$. This shows that increasing the number of particles $K$ beyond the training value can make inference more robust.

\textbf{Ablation of SWB.} We ablate two design choices of SWB. First, we test whether glimpse encoding is necessary for good inference. Second, we test whether file-slot matching indeed requires a summary representation of all the files. We report the results of ablation of SWB in Appendix~\ref{ax:ablation}.

\subsection{Belief-Based Control}
\begin{table}[t]
\centering
\scalebox{0.9}{
\begin{tabular}{@{}lrr@{}}
\toprule
\multicolumn{1}{l}{Model}                         & 2D Maze & 3D Food Chase                                            \\ \midrule
\multicolumn{1}{l|}{Random Policy} &   $-5.69 \pm 2.13$      & $-3.39 \pm 5.58$ \\
\multicolumn{1}{l|}{AESMC $K=10$}  &         $-3.52 \pm 1.23$& $3.71 \pm 8.32$                                                         \\
\multicolumn{1}{l|}{AESMC $K=30$}  &         $-2.37 \pm 2.58$& $5.10 \pm 7.83$                                                         \\
\multicolumn{1}{l|}{SWB $K=1$}       &  $-2.71 \pm 1.53$        & $8.69 \pm 9.81$                                        \\
\multicolumn{1}{l|}{SWB $K=10$}       &         $-2.15 \pm 2.02$ & $10.88 \pm 9.90 $                                           \\
\multicolumn{1}{l|}{SWB $K=30$}    &         $\mathbf{-1.32 \pm 2.30}$& $\mathbf{12.13 \pm 8.59} $                                           \\
\bottomrule
\end{tabular}}
\caption{Performance of Monte Carlo planning on 2D Maze and 3D Food Chase using SWB and AESMC as the world model with different number of particles $K$ used to maintain belief.}
\label{tab:planner}
\vspace{-6mm}
\end{table}

We test how structured belief improves performance in reinforcement learning and planning. We experiment on three games: 2D Branching Sprites game, 2D Maze game and the 3D Food Chase game. We evaluate the performance in terms of average total reward.

\subsubsection{Reinforcement Learning}
For reinforcement learning, we train the policy using A2C that has access to the belief states provided by a world model. We compare the world models: \textit{i)} AESMC, \textit{ii)} non-belief version of SWB i.e. SWB $K=1$, \textit{iii)} SWB with $K=3$ and \textit{iv)} SWB with $K=10$. For SWB, we pre-train the world model with a random exploration policy. By doing this the world model becomes task agnostic. For AESMC, we train the world model jointly with the reward loss as done in DVRL \cite{dvrl}. We will refer to this agent simply as DVRL. Note that by doing this, we are comparing against a stronger baseline with a task-aware world model.

In Fig.~\ref{fig:a2c-all}, we first analyse the performance of DVRL by training the agent on the fully and partially observable version of the 2D Branching Sprites game. We note that even though the agent learns well in the fully-observable game, it fails in the partially-observable version. This suggests that even when AESMC receives gradients from the reward loss, it fails to extract the information about invisible objects required for performing well in the game. Next, in all three games, we show that even a non-belief version of SWB significantly improves agent performance compared to the unstructured belief state of AESMC. This shows the benefit of structured representation which is enabled by our inductive bias of object permanence. In all three games, as we increase the number of particles from 1 to 10, the performance improves further. We hypothesize that this gain results from better modeling of density over the true states of the objects as shown in Fig.~\ref{fig:all_nlls_nospawn} and Fig.~\ref{fig:mota-metrics-analysis}.

\subsubsection{Planning}

To investigate how SWB can improve planning, we use a simple planning algorithm that simulates future roll-outs using random actions up to a fixed depth and predicts an average discounted sum of rewards. We then pick and execute the action that led to the largest average reward. We evaluate the following world models: AESMC and SWB with $K=1, 10$ and 30. By varying $K$, we study the gains of performing planning starting from multiple hypotheses. For fair comparison, we keep the number of simulated futures same for all $K$. In Table \ref{tab:planner}, we observe that AESMC achieves lower score than ours due to poorer generation. Furthermore, having more particles in SWB helps performance. We hypothesize that these are due to the future generation accuracy evaluated in Fig.~\ref{fig:all_nlls_nospawn}.

\subsection{Belief-Based Supervised Reasoning}
To investigate the benefits of providing belief as input for supervised learning, we re-purpose the 2D Maze environment and assign a value to each object color. We also add more than one goal cells. The task is to predict the total sum of all object values present on the goal states. For this, a network takes the belief as input and learns to predicts the sum using a cross-entropy loss. We do not back-propagate the gradients to the world model. In Table \ref{tab:sum_task}, we observe the benefits of structured representation to unstructured belief provided by AESMC. We also observe gains when the number of particles are increased.

\begin{table}[t]
\centering
\scalebox{0.9}{\begin{tabular}{@{}lr@{\hspace{2mm}}r@{\hspace{2mm}}r@{\hspace{2mm}}r@{}}
\toprule
\multicolumn{1}{l|}{Model}  & $K=1$ & $K=3$ & $K=5$  & $K=10$\\ \midrule
\multicolumn{1}{l|}{AESMC} &  $29.3 \pm 2.1$ & $37.6 \pm 3.0$  & $47.2 \pm 1.2$  & $54.2 \pm 5.2$\\
\multicolumn{1}{l|}{SWB}  &  $\mathbf{30.1 \pm 2.4}$ & $\mathbf{43.1 \pm 2.2}$ & $\mathbf{60.8 \pm 3.2}$ & $\mathbf{63.4 \pm 2.6}$\\ 
\bottomrule
\end{tabular}}
\caption{Comparison of accuracy using SWB and AESMC on the supervised reasoning task for different number of particles $K$.}
\label{tab:sum_task}
\vspace{-4mm}
\end{table}
\begin{figure}[t]
    \vspace{0mm}
    \centering
    \includegraphics[width=0.485\textwidth]{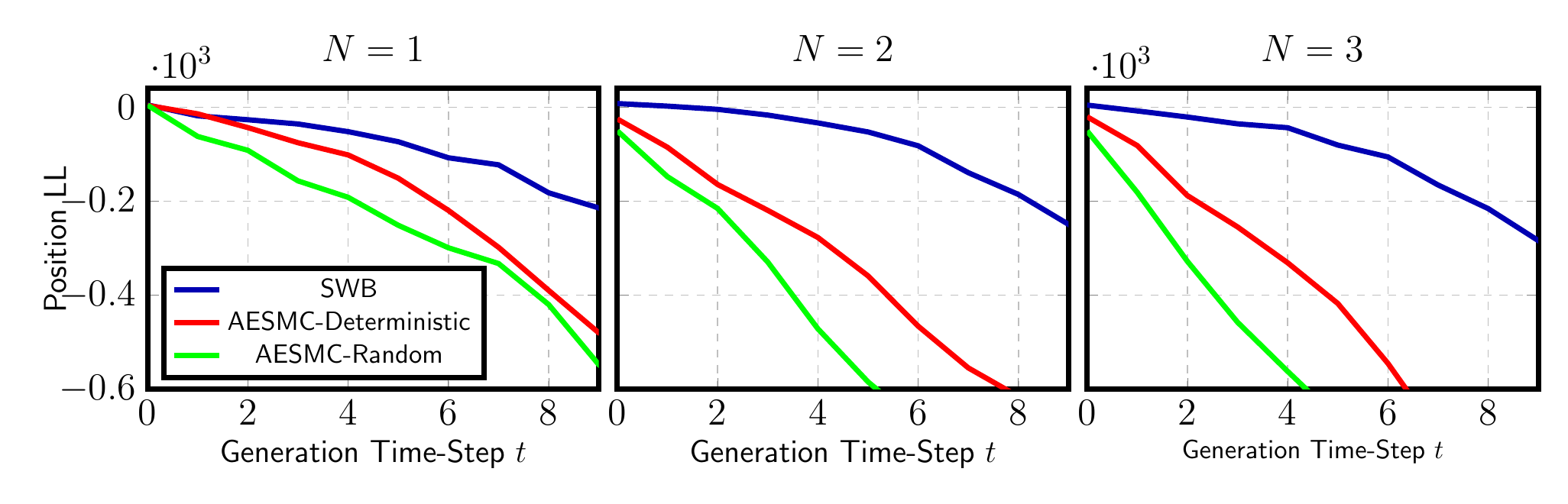}
    \vspace{-8mm}
    \caption{Position log-likelihood on 2D Bouncing Sprites comparing SWB and AESMC on varying number of objects ($N$) and effect of randomness in environment dynamics on AESMC.}
    \label{fig:AESMC-Analysis}
    \vspace{-3mm}
\end{figure}


\subsection{Analysis of AESMC}
We study AESMC along \textit{i)} number of objects in the scene and \textit{ii)} whether the object dynamics are random or deterministic. We provide details about the experimental setup in Appendix \ref{ax:aesmc-analysis}. We note that as the number of objects increase, performance of SWB remains robust. However, AESMC deteriorates due to lack of the structured dynamics modeling via objects and modeling the scene as a single encoding vector.

\section{Conclusion}
We presented a representation learning model combining the benefits of object-centric representation and belief states. Our experiment results on video modeling, RL, planning, and supervised reasoning show that this combination is synergetic, as belief representation makes object-centric representation error-tolerant while the object-centric representation makes the belief more expressive and structured. It would be interesting to extend this work to agent learning in complex 3D environments like the VizDoom environment.

\section*{Acknowledgements}

This work was supported by Electronics and Telecommunications Research Institute (ETRI) grant funded by the Korean government [21ZR1100, A Study of Hyper-Connected Thinking Internet Technology by autonomous connecting, controlling and evolving ways]. The authors would like to thank anonymous reviewers for constructive comments.

\bibliography{refs}
\bibliographystyle{icml2021}

\FloatBarrier
\begin{appendices}
\onecolumn
\section{Additional Results}
\subsection{Ablation of SWB}
\label{ax:ablation}
We ablate two design choices of SWB. First, we investigate whether it is necessary for an object file to look at other files when computing the match probability between the file and a slot. In Table~\ref{tab:ablations} (in appendix), we note that this interaction summary is necessary. In its absence, the number of MOT-A errors increase pointing to poor performance in matching files to slots correctly. In the second ablation, the slots matched to the files were directly used to update the state of a file without using an encoding of the glimpse region. We find the performance to be similar to the model that uses the glimpse. However, when training the world model in the 3D environment, we found the glimpse encoding to lead to notably more stable convergence of the model.
\begin{table*}[h!]
\small
\centering
\begin{tabular}{@{}llllllllll@{}}
\toprule
                         & \multicolumn{2}{l}{Tracking}        & \multicolumn{5}{l}{MOT-A}                 & \multicolumn{2}{l}{Generation} \\ \midrule
Model                    & Position LL & Segment LL & FP   & Miss & Switch & Migrate & Ascend & Position LL    & Segment LL    \\ \midrule
SWB                      & -53.28                 & 16553      & 3.24 & 0.27 & 0.07   & 0.00    & 0.01   & -184.1         & 15797         \\
SWB No-Match-Interact & -83.92                 & 13481      & 6.01 & 17.3 & 0.81   & 0.78    & 0.83   & -194.0         & 7662          \\
SWB No-Glimpse           & -54.26                 & 16416      & 1.30 & 1.06 & 0.26   & 0.20    & 0.35   & -192.3         & 15716         \\\bottomrule
\end{tabular}
\caption{Ablation of File-Slot Matching and Inference in SWB in 2D Branching Sprites. All models were trained with $K=10$. In \textit{SWB No-Match-Interaction}, we omit the file-interaction step that is performed as the first step of file-slot matching. In \textit{SWB No-Glimpse}, the slots that were matched to the files were directly used to update the file without using a glimpse encoding.}
\label{tab:ablations}
\end{table*}
\subsection{Qualitative Results}
\begin{figure}[h!]
    \centering
    \includegraphics[width=1.0\textwidth]{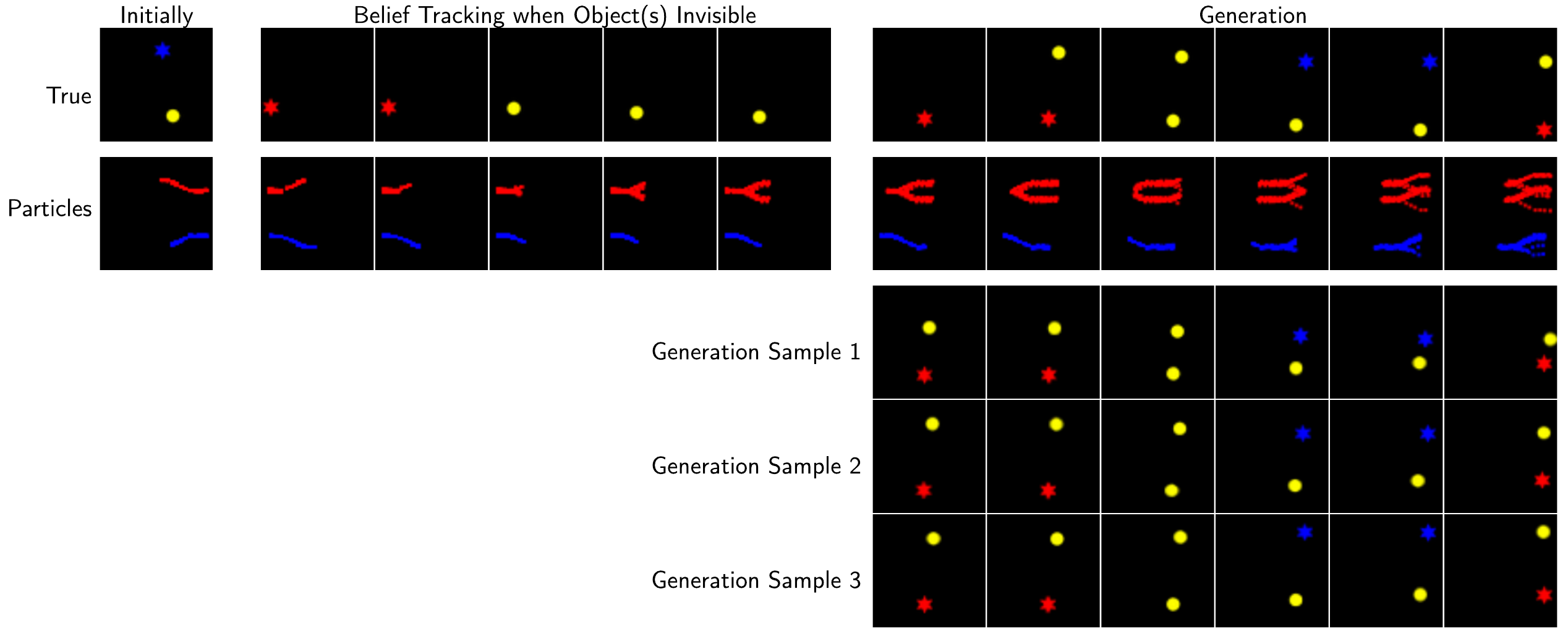}
    \caption{\textbf{Belief Tracking and Generation Samples in 2D Branching Sprites.} The aim of these qualitative results is to show generated object appearances along with object trajectories. In the top row, we show the observations. In the left block we show the object files assigned to each object. Then during belief tracking, some objects can become invisible leading to multi-modal belief. Lastly, taking the multi-modal belief, we perform generation. On the right, we see the generated object positions. We also show generated samples of object appearances. In all three samples, we see that the object appearances match what the true object appearances as shown in the top row.
    }
    \label{fig:qual-2d-gensamples-branching-sprites-ax}
\end{figure}
\begin{figure}[h!]
    \centering
    \includegraphics[width=0.9\textwidth]{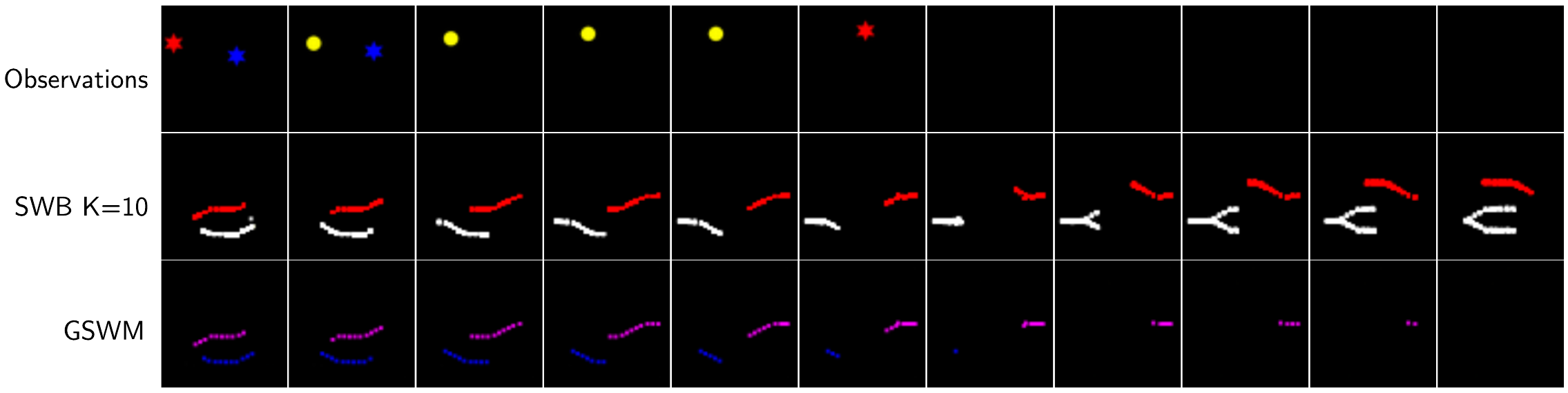}
    \caption{\textbf{Comparison of Belief Tracking in SWB and GSWM.} The top row shows the observations, the middle row shows position particles maintained by SWB with $K=10$ and the bottom row shows the object positions maintained by GSWM. Since GSWM deletes object files for invisible objects due to lack of object permanence, we see that when objects become invisible, there is no belief in GSWM.
    }
    \label{fig:qual-2d-tracksamples-branching-sprites-ax}
\end{figure}
\begin{figure}[h!]
    \centering
    \includegraphics[width=0.9\textwidth]{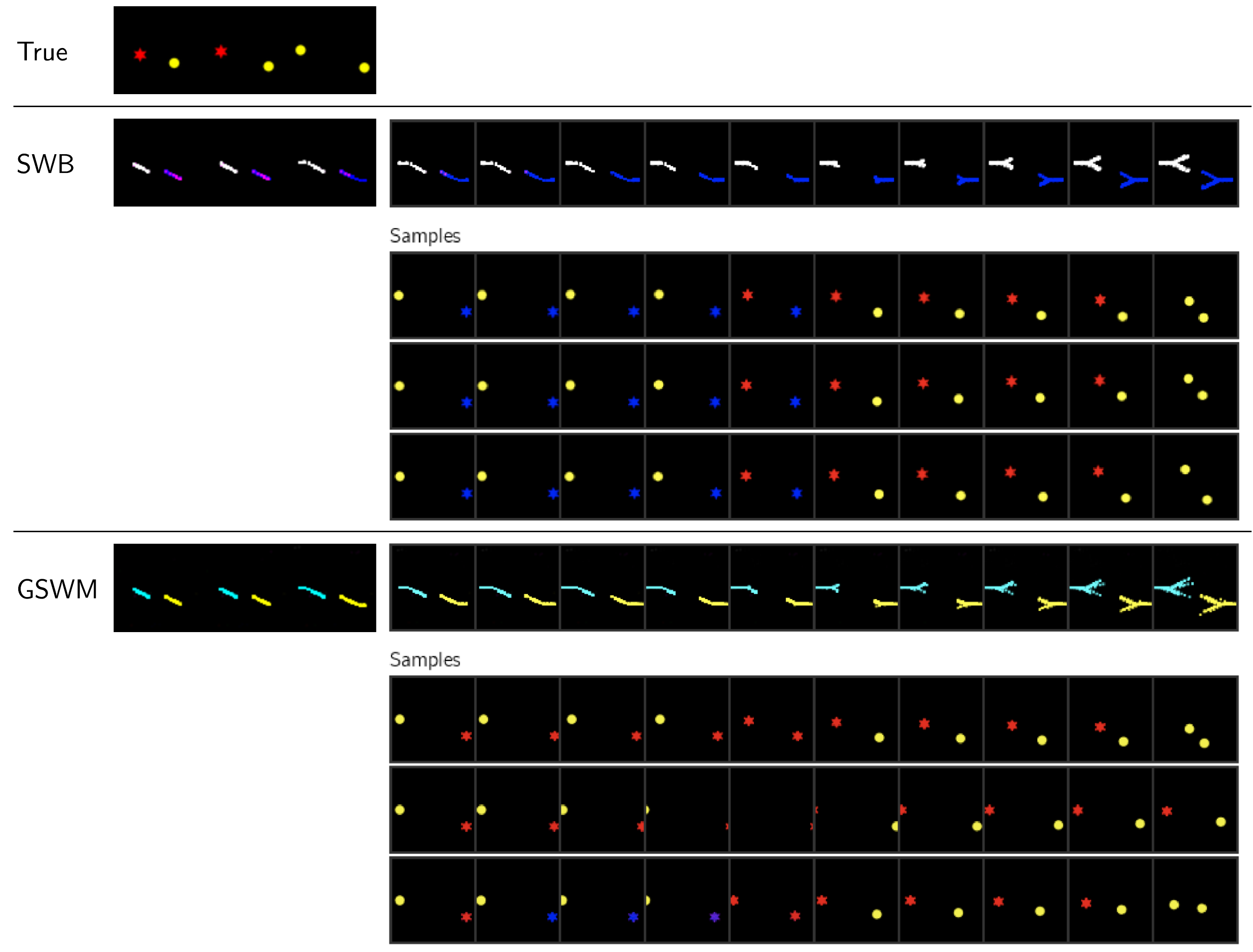}
    \caption{\textbf{Comparison of Generation in SWB and GSWM.} On the left, we show the conditioning observations and the scene representation maintained by the models. These are followed by generations paths obtained by sampling 10 samples of future trajectories. Because GSWM deletes object files for invisible objects, we made sure that the objects were visible in the conditioning period for fair comparison in this figure. We note that the generated samples from GSWM are more inaccurate.
    }
    \label{fig:qual-2d-gensamples-branching-sprites-ax}
\end{figure}
\begin{figure}[h]
    \centering
    \includegraphics[width=1.0\textwidth]{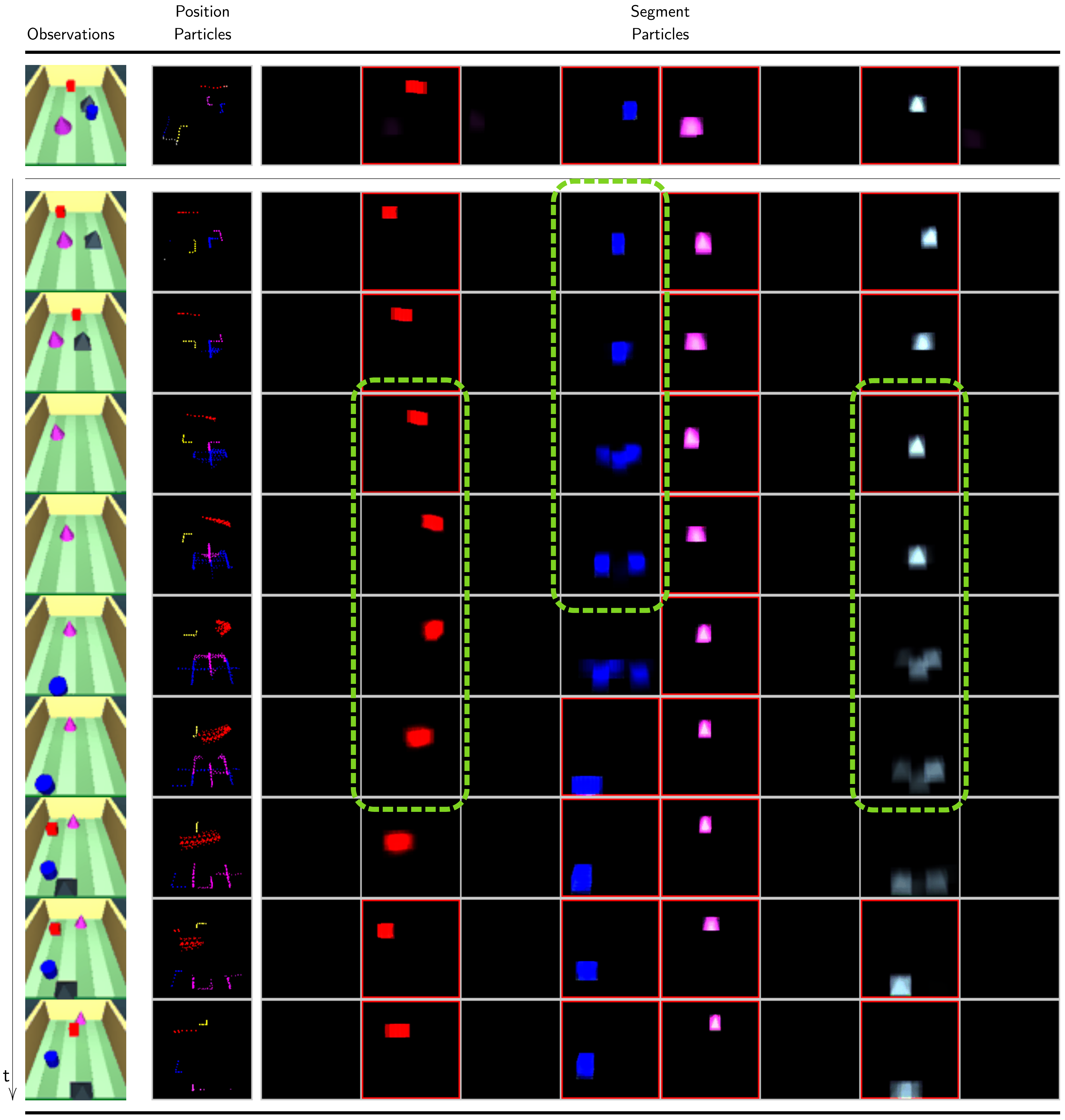}
    \caption{\textbf{Belief Tracking 3D Food Chase Game.} Each row corresponds to one time-step and we show time-steps at increments of 4. The left-most column shows the observations shown to the model. The second column shows the position particles maintained by the model. The remaining columns show the rendering of the object segments for each of the 8 object files. In these, we make the cell-border red when the belief infers the object file to be \textit{visible} in that time-step. For each object file, we super-impose the segments for all particles by averaging over $K$. We note that the model can maintain belief states over the invisible objects. The dashed green cells highlight when the objects are invisible and the model is maintaining a belief of diverse and plausible states for those object files. Also note that when the objects become visible again, the correct object files can re-attach to track them.}
    \label{fig:qual-3d-ax}
\end{figure}
\begin{figure}[h!]
    \centering
    \includegraphics[width=0.7\textwidth]{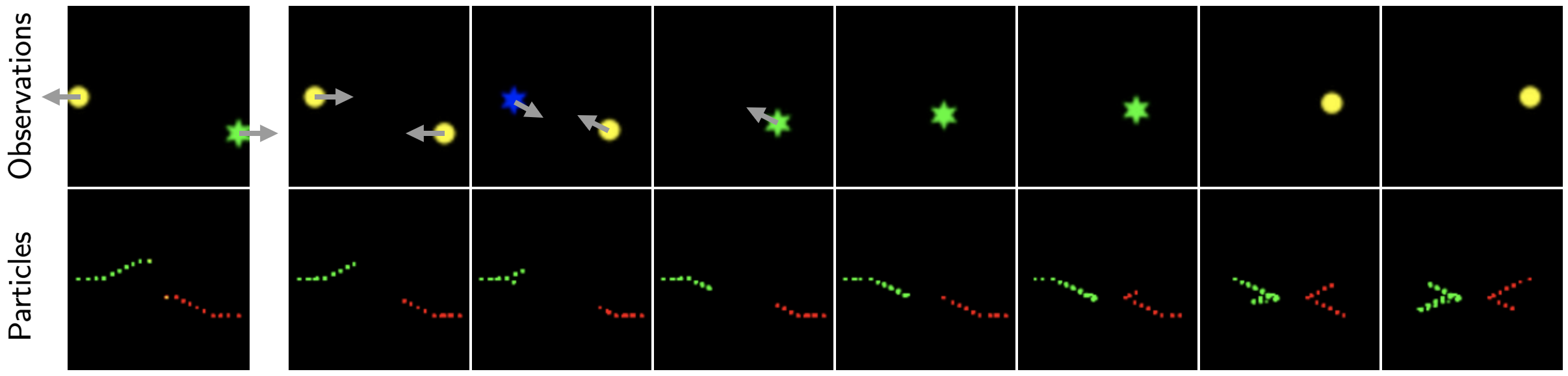}
    \caption{\textbf{Demonstration of interaction between objects in SWB.}  In the observation, we see that one object becomes invisible while the other remains visible. The grey arrows show the direction of motion of the true objects. We see that invisible object file (shown as green colored particles) can simulate the collision with the visible object file (shown as red colored particles) and maintain correct belief about the invisible object.
    }
    \label{fig:interaction-bw-particles}
\end{figure}
\begin{figure}[h!]
    \centering
    \includegraphics[width=1.0\textwidth]{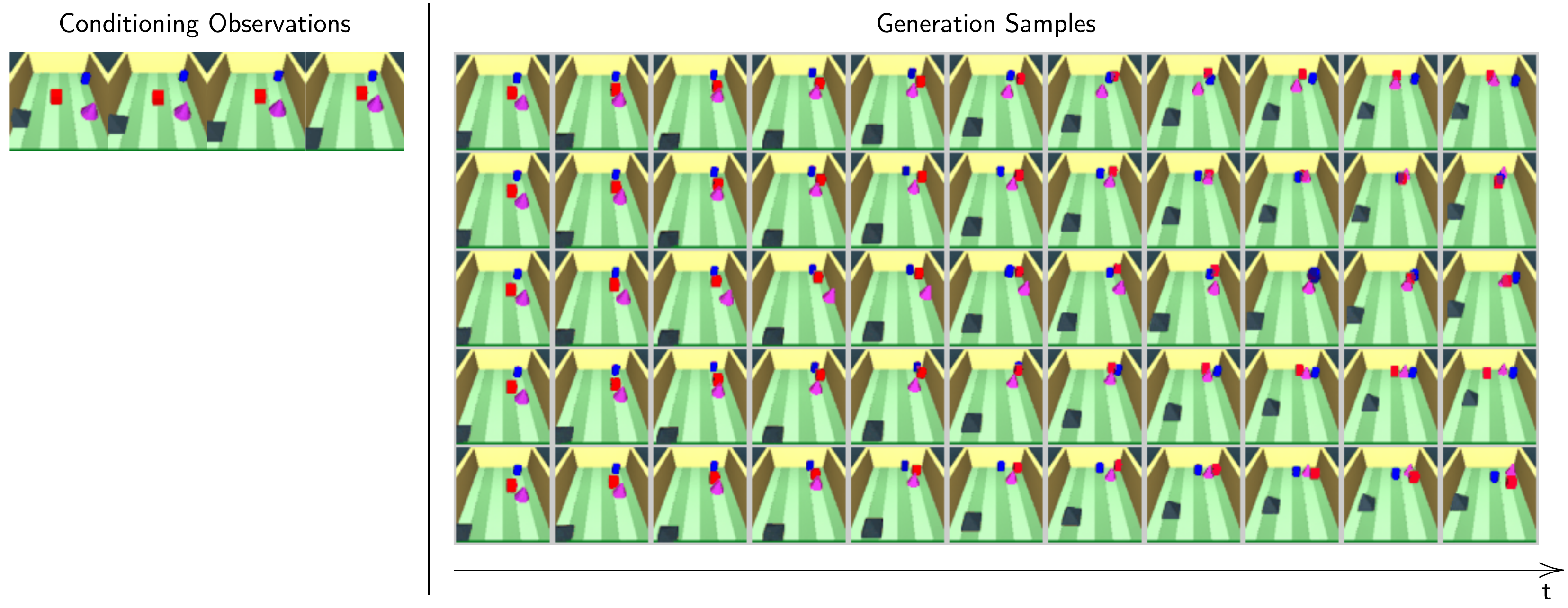}
    \caption{\textbf{Qualitative Results of Generation in SWB in 3D Food Chase Game.} On the left we show the conditioning observations. On the right, each row is one sample. Each column corresponds to time-steps at intervals of 2. We note that the generations are plausible and diverse.}
    \label{fig:qual-3d-gen-ax}
\end{figure}
\begin{figure}[h!]
    \centering
    \includegraphics[width=0.8\textwidth]{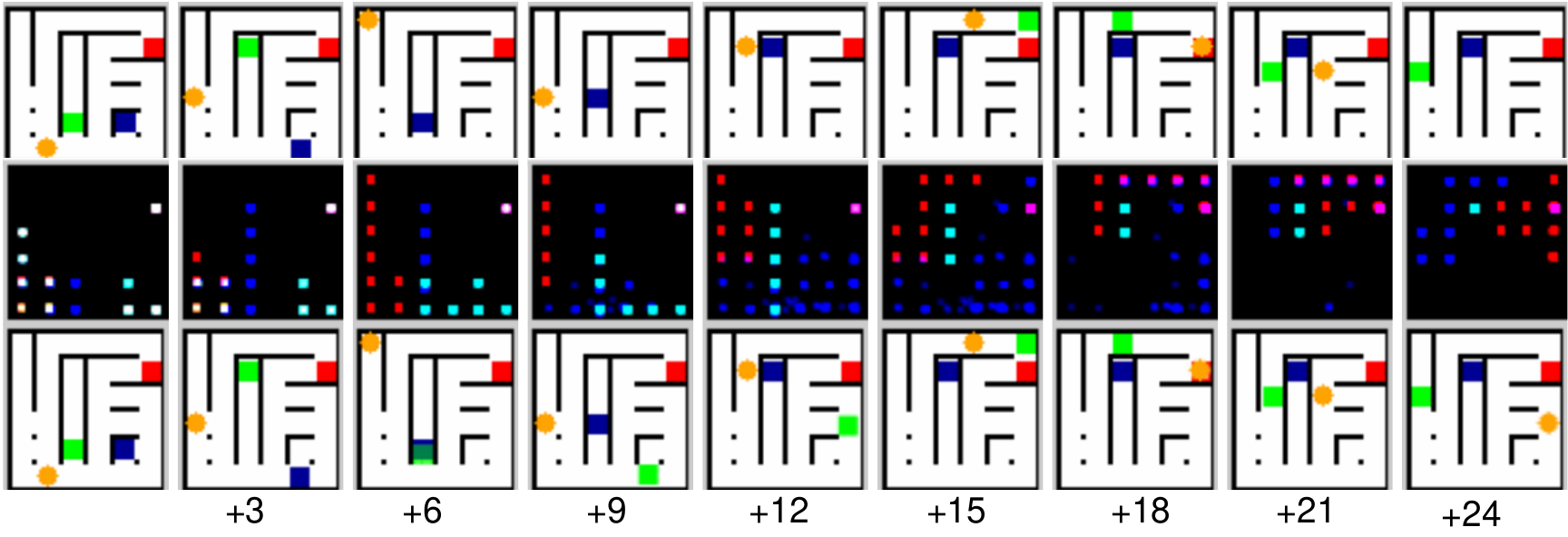}
    \caption{\textbf{Belief Tracking in 2D Maze.} In the top row, we show the observations. Middle row visualizes 8 most recent position states of each active object file. The color of the file denotes ID. The bottom row shows the reconstruction obtained from particle with index $k=1$ at each time-step. We note in \red{+3} that the green object is tracked by the blue object file. The green object becomes invisible in \red{+6} and from \red{+6} to \red{+15}, the blue object file is able to track and maintain plausible particles for it (shown in bottom row). 
    }
    \label{fig:qual-2d-maze-ax}
\end{figure}
\FloatBarrier
\subsection{Additional Quantitative Results}
\begin{table}[h!]
\centering
\begin{tabular}{@{}lrr@{}}
\toprule
Model & Position LL & Segment LL \\ \midrule
SWB $K=10$ $N=8$   & -39.18      & 23306      \\
SWB $K=20$ $N=8$    & -23.31      & 23353      \\
SWB $K=30$ $N=8$    & $\mathbf{-15.74}$      & $\mathbf{23365}$      \\ \midrule
GSWM  & -993.3      & 22457      \\ \bottomrule
\end{tabular}
\caption{\textbf{Generalization to more number of objects and object files in SWB belief tracking.} We evaluate SWB trained on 2D Branching Sprites (No-Spawn-2) with 4 object files and test on 2D Branching Sprites (Spawn-4) using 8 object files. We report the average log-likelihood of ground truth position and ground truth object segments evaluated using kernel density estimation on SWB particles. We note the benefits of using higher number of particles in obtaining improved density modeling. Furthermore, we show that we outperform our baseline GSWM primarily because it does not have object permanence and does not maintain files for invisible objects.}
\end{table}
\section{Derivations}
In this section, we provide the derivation of the ELBO training objective and the derivation of the belief update step.
\subsection{Derivation of the ELBO}
Our objective is to maximize the following marginal log-likelihood over the observations $\bx_{1:T}$, i.e.~$\log p_\ta(\bx_{1:T})$ under our latent variable model $p_\ta(\bx_{1:T}, \bs_{1:T})$. Here, integrating over the high-dimensional latent $\bs_{1:T}$ is intractable. Hence, we resort to AESMC and approximate this via an ELBO lower bound. To do this, we first re-write the objective $\log p_\ta(\bx_{1:T})$ as:
\begin{align}
    \log p_\ta(\bx_{1:T}) &= \log \sum_{k=1}^K \frac{1}{K} p_\ta(\bx_{1:T}|\bs_{0}^k = \texttt{null}).
\end{align}
To evaluate this objective, we consider the following general expression and then proceed to approximate it.
\begin{align}
    \log \sum_{k=1}^K \bar{w}_{t-1}^k p_\ta(\bx_{t:T}|\bs_{t-1}^k). \label{elbo:start}
\end{align}
Here, $\bar{w}_{t-1}^k$ are the normalized weights of the particles based on the observations seen thus far (bar denotes normalization), $\bs_{t-1}^k$ are the particle states for the previous time-step and $\bx_{t:T}$ denotes all the observations starting from the current time-step $t$ up to the end of the training horizon $T$.
To evaluate this objective, we first perform re-sampling of the particles $\bs_{t-1}^{1:K}$ using the normalized weights $\bar{w}_t^{1:K}$. This is done because some particles may have vanishingly small particle weights that explain poorly the observations received so far. Hence, by performing resampling, we eliminate those particles from our particle pool. 

In resampling, we sample a set of ancestor indices $a_{t-1}^k \sim q_\text{R}(\cdot|\bar{w}_{t-1}^{1:K})$ from a categorical distribution. In our work, we use soft-resampling as the sampling distribution $q_\text{R}$ for these ancestor indices and we can write \eqref{elbo:start} as follows.
\begin{align}
    \log \eE_{a_{t-1}^{k} \sim q_\text{R}(\cdot|\bar{w}_{t-1}^{1:K})} \sum_{k=1}^K \tilde{w}_{t-1}^{a_{t-1}^k} p_\ta(\bx_{t:T}|\bs_{t-1}^{a_{t-1}^k}). \label{elbo:softresample}
\end{align}
where
\begin{align}
    q_\text{R}(k|\bar{w}_{t-1}^{1:K}) &= \alpha \bar{w}_{t-1}^k + (1 - \alpha)\frac{1}{K}, \\
     a_{t-1}^{k} &\sim q_\text{R}(\cdot|\bar{w}_{t-1}^{1:K}) \quad\forall k = 1, 2, \ldots, K,\\
    \tilde{w}_\tmo^k &\leftarrow \frac{\bar{w}_\tmo^{k}}{\alpha \bar{w}_\tmo^{k} + (1 - \alpha)\frac{1}{K}}.
\end{align}
where $\alpha$ is a trade-off hyper-parameter. It can be verified that \eqref{elbo:softresample} is an unbiased estimator of \eqref{elbo:start}. Next, we apply Jensen's inequality on \eqref{elbo:softresample} to get the following.
\begin{align}
    \log & \eE_{a_{t-1}^{k} \sim q_\text{R}(\cdot|\bar{w}_{t-1}^{1:K})} \sum_{k=1}^K \tilde{w}_{t-1}^{a_{t-1}^k} p_\ta(\bx_{t:T}|\bs_{t-1}^{a_{t-1}^k}) \\
     &\geq  \eE_{a_{t-1}^{k} \sim q_\text{R}(\cdot|\bar{w}_{t-1}^{1:K})} \log \sum_{k=1}^K \tilde{w}_{t-1}^{a_{t-1}^k} p_\ta(\bx_{t:T}|\bs_{t-1}^{a_{t-1}^k}) \\
     &=  \eE_{a_{t-1}^{k} \sim q_\text{R}(\cdot|\bar{w}_{t-1}^{1:K})} \log \sum_{k=1}^K \tilde{w}_{t-1}^{a_{t-1}^k} \int p_\ta(\bx_t|\bs_{t}^k) p_\ta(\bs_t^k|\bs_{t-1}^{a_{t-1}^k}) p_\ta(\bx_{t+1:T}|\bs_{t}^{k}) \,d \bs_t^k\\
     &=  \eE_{a_{t-1}^{k} \sim q_\text{R}(\cdot|\bar{w}_{t-1}^{1:K})} \log \sum_{k=1}^K \tilde{w}_{t-1}^{a_{t-1}^k} \eE_{\bs_t^{k}\sim q_{\ta,\phi}(\bs_{1}|\bs_{t-1}^{a_{t-1}^k}, \bx_t)} p_\ta(\bx_t|\bs_{t}^k) \dfrac{p_\ta(\bs_t^k|\bs_{t-1}^{a_{t-1}^k})}{q_{\ta, \phi}(\bs_t^k|\bs_{t-1}^{a_{t-1}^k}, \bx_t)} p_\ta(\bx_{t+1:T}|\bs_{t}^{k}) \\
     &\geq  \eE_{a_{t-1}^{k} \sim q_\text{R}(\cdot|\bar{w}_{t-1}^{1:K})} \eE_{\bs_t^{k}\sim q_{\ta,\phi}(\bs_{1}|\bs_{t-1}^{a_{t-1}^k}, \bx_t)} \log \sum_{k=1}^K {\ubm{\tilde{w}_{t-1}^{a_{t-1}^k}  p_\ta(\bx_t|\bs_{t}^k) \dfrac{p_\ta(\bs_t^k|\bs_{t-1}^{a_{t-1}^k})}{q_{\ta, \phi}(\bs_t^k|\bs_{t-1}^{a_{t-1}^k}, \bx_t)}}{{w_t^k}}} p_\ta(\bx_{t+1:T}|\bs_{t}^{k})\\
     &=  \eE_{a_{t-1}^{k} \sim q_\text{R}(\cdot|\bar{w}_{t-1}^{1:K})} \eE_{\bs_t^{k}\sim q_{\ta,\phi}(\bs_{1}|\bs_{t-1}^{a_{t-1}^k}, \bx_t)} \log \sum_{k=1}^K w_t^k p_\ta(\bx_{t+1:T}|\bs_{t}^{k}) \\
     &=  \eE_{a_{t-1}^{k} \sim q_\text{R}(\cdot|\bar{w}_{t-1}^{1:K})} \eE_{\bs_t^{k}\sim q_{\ta,\phi}(\bs_{1}|\bs_{t-1}^{a_{t-1}^k}, \bx_t)} \log \left(\sum_{j=1}^K w_t^j\right)\sum_{k=1}^K \frac{w_t^k}{\sum_{j=1}^K w_t^j} p_\ta(\bx_{t+1:T}|\bs_{t}^{k}) \\
     &=  \eE_{a_{t-1}^{k} \sim q_\text{R}(\cdot|\bar{w}_{t-1}^{1:K})} \eE_{\bs_t^{k}\sim q_{\ta,\phi}(\bs_{1}|\bs_{t-1}^{a_{t-1}^k}, \bx_t)} \log \left(\sum_{k=1}^K w_t^k\right) \\&\quad\quad\quad+ \eE_{a_{t-1}^{k} \sim q_\text{R}(\cdot|\bar{w}_{t-1}^{1:K})} \eE_{\bs_t^{k}\sim q_{\ta,\phi}(\bs_{1}|\bs_{t-1}^{a_{t-1}^k}, \bx_t)} \log \sum_{k=1}^K \frac{w_t^k}{\sum_{j=1}^K w_t^j} p_\ta(\bx_{t+1:T}|\bs_{t}^{k})\\
     &=  \eE_{a_{t-1}^{k} \sim q_\text{R}(\cdot|\bar{w}_{t-1}^{1:K})} \eE_{\bs_t^{k}\sim q_{\ta,\phi}(\bs_{1}|\bs_{t-1}^{a_{t-1}^k}, \bx_t)} \log \left(\sum_{k=1}^K w_t^k\right) \\&\quad\quad\quad+ \eE_{a_{t-1}^{k} \sim q_\text{R}(\cdot|\bar{w}_{t-1}^{1:K})} \eE_{\bs_t^{k}\sim q_{\ta,\phi}(\bs_{1}|\bs_{t-1}^{a_{t-1}^k}, \bx_t)} \log \sum_{k=1}^K \bar{w}_t^k p_\ta(\bx_{t+1:T}|\bs_{t}^{k}).
\end{align}
Therefore, we have
\begin{align}
\label{ineq:smc}
\begin{split}
    \log \sum_{k=1}^K \bar{w}_{t-1}^k p_\ta(\bx_{t:T}|\bs_{t-1}^k) 
    &\geq  \eE_{a_{t-1}^{k} \sim q_\text{R}(\cdot|\bar{w}_{t-1}^{1:K})} \eE_{\bs_t^{k}\sim q_{\ta,\phi}(\bs_{1}|\bs_{t-1}^{a_{t-1}^k}, \bx_t)} 
    \ubm{\log \left(\sum_{k=1}^K w_t^k\right)}{\text{Term I}}  
    \\&\quad\quad\quad+ \eE_{a_{t-1}^{k} \sim q_\text{R}(\cdot|\bar{w}_{t-1}^{1:K})} \eE_{\bs_t^{k}\sim q_{\ta,\phi}(\bs_{1}|\bs_{t-1}^{a_{t-1}^k}, \bx_t)} \ubm{\log \sum_{k=1}^K \bar{w}_t^k p_\ta(\bx_{t+1:T}|\bs_{t}^{k})}{\text{Term II}}.
    \end{split}
\end{align}
Here, the term II can be expanded recursively by re-applying the inequality \eqref{ineq:smc}. Applying this inequality recursively on our original objective $\log p(\bx_{1:T})$, we get the following ELBO.
\begin{align}
   \log p(\bx_{1:T}) \geq \sum_{t=1}^T \eE_{a_{1:t-1}^{k}, \bs_{1:t}^{k}} \log \left(\sum_{k=1}^K w_t^k\right).
\end{align}\qed
\subsection{Decomposing the Weight Term $w_t^k$}
In the above derivation, we had 
\begin{align}
    w_t^k &= \tilde{w}_{t-1}^{a_{t-1}^k}  p_\ta(\bx_t|\bs_{t}^k) \dfrac{p_\ta(\bs_t^k|\bs_{t-1}^{a_{t-1}^k})}{q_{\ta, \phi}(\bs_t^k|\bs_{t-1}^{a_{t-1}^k}, \bx_t)} \label{eq:ax:weightupdate}
\end{align}
Based on the generative model of SWB, we have
\begin{align}
p_\ta(\bs_{t,n} | \bs_{t-1}) &= \ub{p_\ta(\bz_{t,n}| \bz_{t-1}, \bh_{t-1})}{Dynamics Prior} \ub{p(i_{t,n}^k)}{ID Prior},\nn\\
&= \ub{p_\ta(\bz_{t,n}^{\text{vis}}| \bz_{t-1}, \bh_{t-1})}{Visibility Prior} \ub{p_\ta(\bz_{t,n}^{\text{obj}}| \bz_{t-1}, \bh_{t-1})}{Object Prior} \ub{p(i_{t,n}^k)}{ID Prior}.\nn
\end{align}
Additionally, for the sampled file-slot matching index $m_{t,n}$, we take a uniform categorical distribution as the prior $p(m_{t,n})$. This stems from the idea that there is no inductive bias to prefer matching one slot over another. Similarly, based on the file inference model of SWB, we have
\begin{align}
q_{\ta,\phi}(\bs_{t,n} | \bs_{t-1}, \bx_t)
&=  \ub{q_{\ta,\phi}(\bz_{t,n}^{\text{obj}}| \bz_{t-1}, \bh_{t-1}, \bx_t)}{Object Inference} \ub{q_{\phi}(i_{t,n}^k | \bz_{t,n}^{\text{vis}}, i_{t-1,n}^k)}{ID Inference}\ub{q_{\phi}(\bz_{t,n}^{\text{vis}}| \bz_{t-1}, \bh_{t-1}, \bx_t)}{Visibility Inference}\ub{q_{\phi}(m_{t,n}^k | \bz_{t-1}, \bh_{t-1}, \bx_t)}{Slot Matching}.\nn
\end{align}
Substituting these into \eqref{eq:ax:weightupdate}, we get
\begin{align}
    w_t^k &= \tilde{w}_{t-1}^{a_{t-1}^k}  p_\ta(\bx_t|\bs_{t}^k) \dfrac{\prod_{n=1}^N{p_\ta(\bz_{t,n}^{\text{vis}}| \bz_{t-1}, \bh_{t-1})} {p_\ta(\bz_{t,n}^{\text{obj}}| \bz_{t-1}, \bh_{t-1})} {p(i_{t,n}^k)} p(m_{t,n}) }{\prod_{n=1}^N{q_{\ta,\phi}(\bz_{t,n}^{\text{obj}}| \bz_{t-1}, \bh_{t-1}, \bx_t)} {q_{\phi}(i_{t,n}^k | \bz_{t,n}^{\text{vis}}, i_{t-1,n}^k)}{q_{\phi}(\bz_{t,n}^{\text{vis}}| \bz_{t-1}, \bh_{t-1}, \bx_t)}{q_{\phi}(m_{t,n}^k | \bz_{t-1}, \bh_{t-1}, \bx_t)}},\\
    &= \tilde{w}_{t-1}^{a_{t-1}^k}  p_\ta(\bx_t|\bs_{t}^k) \prod_{n=1}^N\dfrac{{p_\ta(\bz_{t,n}^{\text{vis}}| \bz_{t-1}, \bh_{t-1})} {p_\ta(\bz_{t,n}^{\text{obj}}| \bz_{t-1}, \bh_{t-1})} {p(i_{t,n}^k)} p(m_{t,n}) }{{q_{\ta,\phi}(\bz_{t,n}^{\text{obj}}| \bz_{t-1}, \bh_{t-1}, \bx_t)} {q_{\phi}(i_{t,n}^k | \bz_{t,n}^{\text{vis}}, i_{t-1,n}^k)}{q_{\phi}(\bz_{t,n}^{\text{vis}}| \bz_{t-1}, \bh_{t-1}, \bx_t)}{q_{\phi}(m_{t,n}^k | \bz_{t-1}, \bh_{t-1}, \bx_t)}},\\
    &= \tilde{w}_{t-1}^{a_{t-1}^k}  p_\ta(\bx_t|\bs_{t}^k) \prod_{n=1}^N 
    \ubm{\dfrac{p_\ta(\bz_{t,n}^{\text{vis}}| \bz_{t-1}, \bh_{t-1})}{q_{\phi}(\bz_{t,n}^{\text{vis}}| \bz_{t-1}, \bh_{t-1}, \bx_t)} \dfrac{p_\ta(\bz_{t,n}^{\text{obj}}| \bz_{t-1}, \bh_{t-1})}{q_{\ta,\phi}(\bz_{t,n}^{\text{obj}}| \bz_{t-1}, \bh_{t-1}, \bx_t)}}{w_{t,n}^{k,\text{file}}}\ubm{\dfrac{p(i_{t,n}^k)}{q_{\phi}(i_{t,n}^k | \bz_{t,n}^{\text{vis}}, i_{t-1,n}^k)}}{w_{t,n}^{k,\text{ID}}} \ubm{\dfrac{p(m_{t,n})}{q_{\phi}(m_{t,n}^k | \bz_{t-1}, \bh_{t-1}, \bx_t)}}{w_{t,n}^{k,\text{match}}}.
\end{align}
\qed
\section{Evaluation Details}
\subsection{Tracking and Generation Metrics}
\label{ax:metrics}
We measure the accuracy of our belief with respect to the ground truth object positions $\by_{t,j}^\text{pos}$ and segments $\by_{t,j}^\text{seg}$, where $j$ is an object. Here, $\by_{t,j}^\text{pos} \in \mathbb{R}^2$ refers to the object coordinates and $\by_{t,j}^\text{seg} \in \mathbb{R}^{C\times W\times H}$ refers to a full-sized empty image with single object rendered on it. Note that true values of these are available even when the object is invisible. We use the particles from the models to perform kernel density estimation of the distribution over the ground truth quantities $\by_{t}$ and we evaluate and report the log-likelihood as follows.
\vspace{-2mm}
\begin{align}
\log p(\by_t) = \log \sum_k w_k \prod_{j} p(\by_{t,j} | \bs_{t,n_j}^k).\nn
\end{align}
where $n_j$ is the object file matched with a ground truth object $j$ at every time-step using the MOT evaluation approach in \cite{milan2016mot16}. If the object file $\bs_{t,n_j}^k$ is not available due to deletion, we assign a random coordinate on the image as the object position and a black image as the object segment. We call these metrics \textit{position log-likelihood} and \textit{segment log-likelihood}. We also compute MOT-A components (such as False Positives or Switches) by counting them individually per particle $k$ and taking weighted average using the particle weights $w_t^k$.

\subsection{Environments}
\label{ax:envs}
\begin{figure}[h!]
    \centering
    \includegraphics[width=0.8\textwidth]{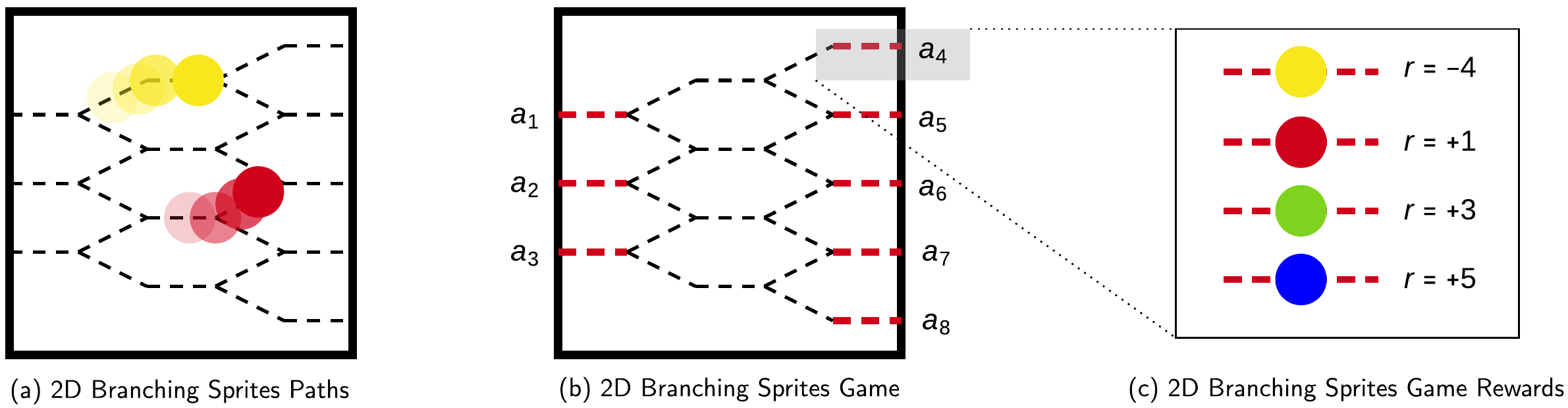}
    \caption{\textbf{2D Branching Sprites Environment Scheme.} In \textit{(a)}, we show the paths which the sprites can take. Note the branching pattern which enables multi-modal position belief to emerge when the objects are invisible. In \textit{(b)}, we turn the environment into a game. The edges highlighted in red can be selected as 8 + 1 actions (1 for not selecting any edge). If an object is on that edge and is also invisible, the agent receives a reward based on the color of the sprite as shown in \textit{(c)}.}
    \label{fig:branching-sprites-game-rules}
\end{figure}

\begin{figure}[h!]
    \vspace{-2mm}
    \centering
    \includegraphics[width=0.6\linewidth]{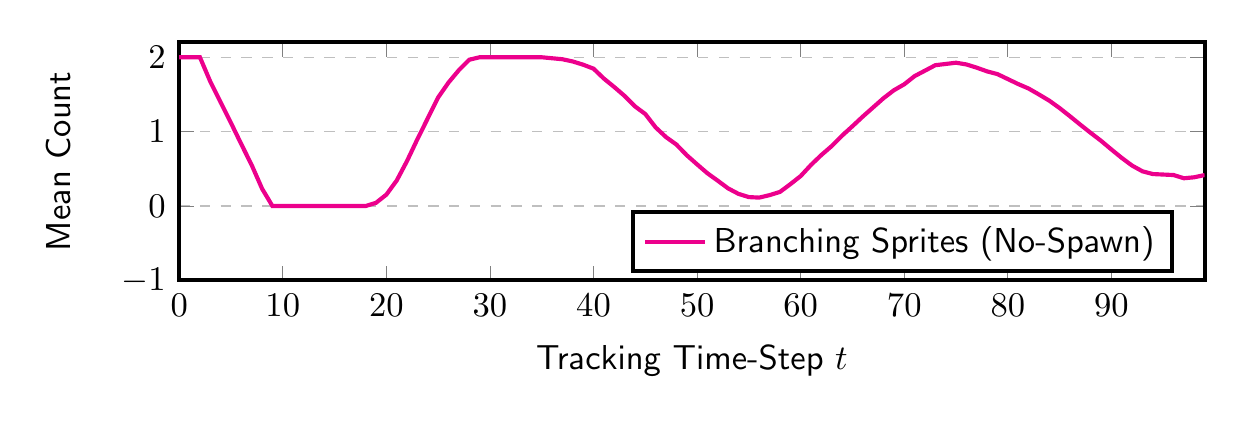}
    \vspace{-5mm}
    \caption{Mean number of objects visible for each time-step in the episode for 2D Branching Sprites (No-Spawn-2). This periodic variation takes place because the duration of invisibility is randomly sampled from $\operatorname{Uniform}(15, 20)$ and duration of visibility is sampled from $\operatorname{Uniform}(15, 30)$. The total number of objects present in the scene is 2.}
    \label{fig:mean-visible-count-branching-sprites-nospawn}
\end{figure}

\begin{figure}[h!]
    \vspace{-2mm}
    \centering
    \includegraphics[width=0.6\linewidth]{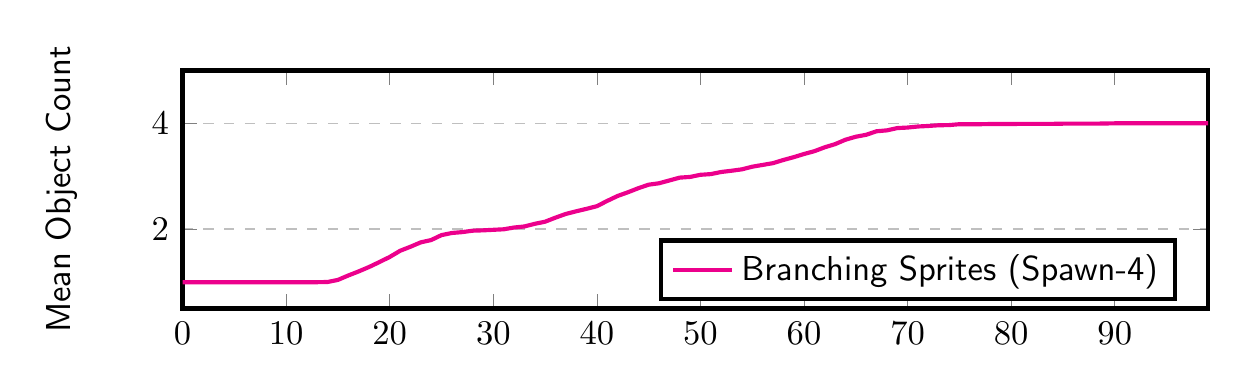}
    \vspace{-5mm}
    \caption{Mean number of objects present in the scene (visible or invisible) in the 2D Branching Sprites (Spawn-4) in which up to 4 objects spawn during the episode.}
    \label{fig:mean-obj-count-2dbs-4-spawner}
\end{figure}

\textit{2D Branching Sprites.} This is a 2D environment with moving sprites. To create a long-occlusion setting, the sprites can disappear for up to 40 time-steps before reappearing. To produce multi-modal position belief during the invisible period, the object paths split recursively with objects randomly taking one branch at every split (see Fig.~\ref{fig:branching-sprites-game-rules}). To produce belief states with appearance change during the invisible period, each sprite is associated with a pair of colors and the color switches periodically every 5 time-steps. We evaluate three versions of this environment: \textit{i) Spawn-$2$:} To evaluate the model in handling new object discovery, we allow upto 2 sprites to spawn during the episode in this version. \textit{ii) Spawn-$4$:} To evaluate the model in handling new object discovery, we allow upto 4 sprites to spawn during the episode in this version. We show the  mean object counts during the episode in Fig.~\ref{fig:mean-obj-count-2dbs-4-spawner}. \textit{iii) No-Spawn-2:} To evaluate the accuracy of belief disentangled from the ability to handle new objects, in this version, the number of sprites remain fixed to two during the episode. We show the mean number of objects visible at each time-step in Fig.~\ref{fig:mean-visible-count-branching-sprites-nospawn}.

\textit{2D Branching Sprites Game.} To test the benefit of our belief for agent learning, we interpret the 2D Branching Sprites environment as a game in which the agent takes an action by selecting a branch on which an invisible object is moving. This requires agent to know which branches are likely to have an object. We make the reward and penalty depend on object color. Hence, the agent also needs to accurately know the object color. The length of invisibility is sampled from range $\operatorname{Uniform}(25, 40)$ and the length of visibility is sampled from range $\operatorname{Uniform}(15, 30)$.

\begin{figure}[h!]
    \centering
    \includegraphics[width=0.7\textwidth]{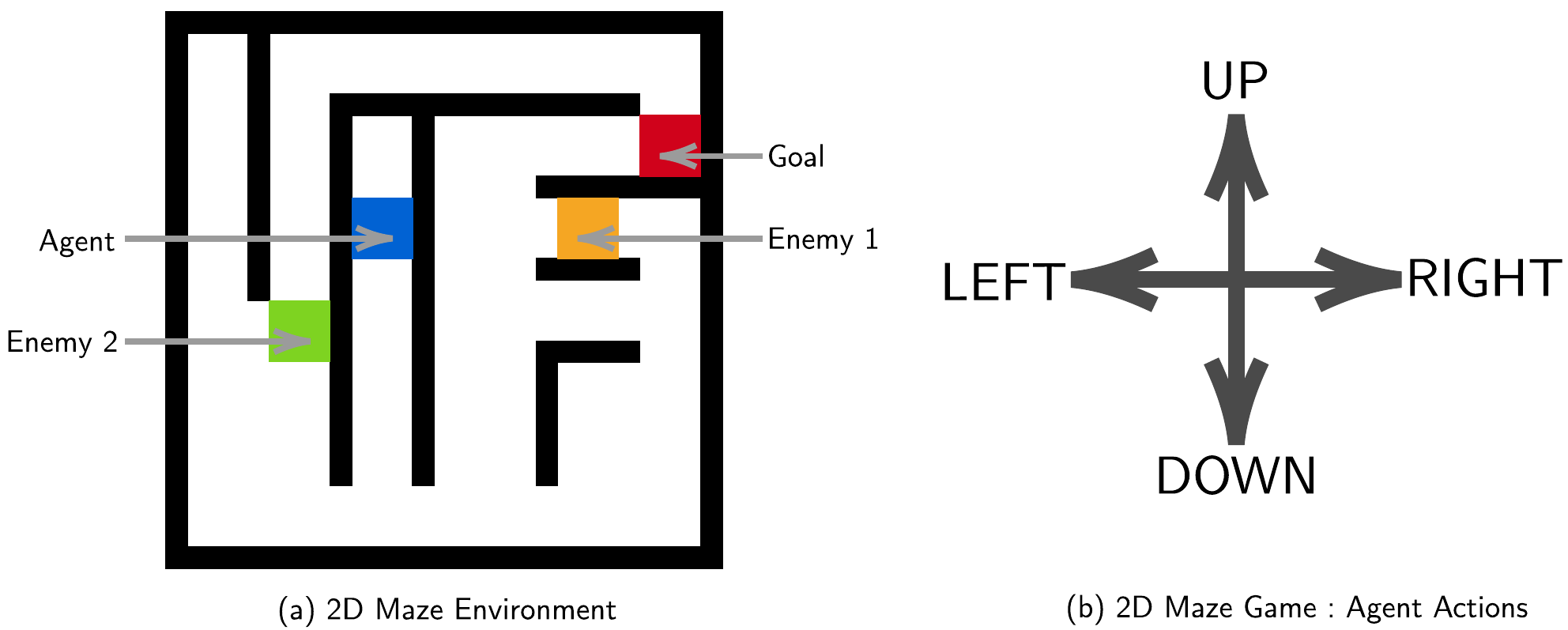}
    \caption{\textbf{2D Maze Game Scheme.} We illustrate the 2D Maze environment. The game comprises of an agent and a goal state which needs to be reached. There are two enemies which need to be avoided.}
    \label{fig:maze-game-rules}
\end{figure}

\textit{2D Maze Game.} We consider a 2D maze environment (see Fig.~\ref{fig:maze-game-rules}) with the blue square as the agent, a red colored goal region and two randomly moving enemy objects. The maze is randomly generated in each episode with the objects navigating through the corridors. The enemies can disappear to create partial observability. The agent receives positive $+1$ reward for reaching the goal, $-1$ reward for hitting an enemy and a small positive reward of $0.025$ for each step it moves to encourage exploration. The data set was created using \cite{gym-maze}.

\textit{3D Food Chase Game.} We use this environment to test our model and our agent in a visually rich 3D game. The environment is a 3D room with the agent as a red cube, food as a blue cylinder and two enemy objects. To enable multi-modal object trajectories, the food and the enemies move along lanes and can randomly change lanes at the intersections. For long invisibility, all objects can disappear for a randomly sampled duration of up to 40 time-steps. The agent receives negative reward for hitting an enemy and positive reward for eating the food. We provide higher reward for eating the food when either the agent or the food is invisible when the food is eaten. The length of invisibility is sampled from range $\operatorname{Uniform}(10, 25)$ and the length of visibility is sampled from range $\operatorname{Uniform}(20, 30)$. The data set was implemented on MuJoCo physics environment \cite{mujoco}.

\begin{figure}[h!]
    \centering
    \includegraphics[width=0.8\textwidth]{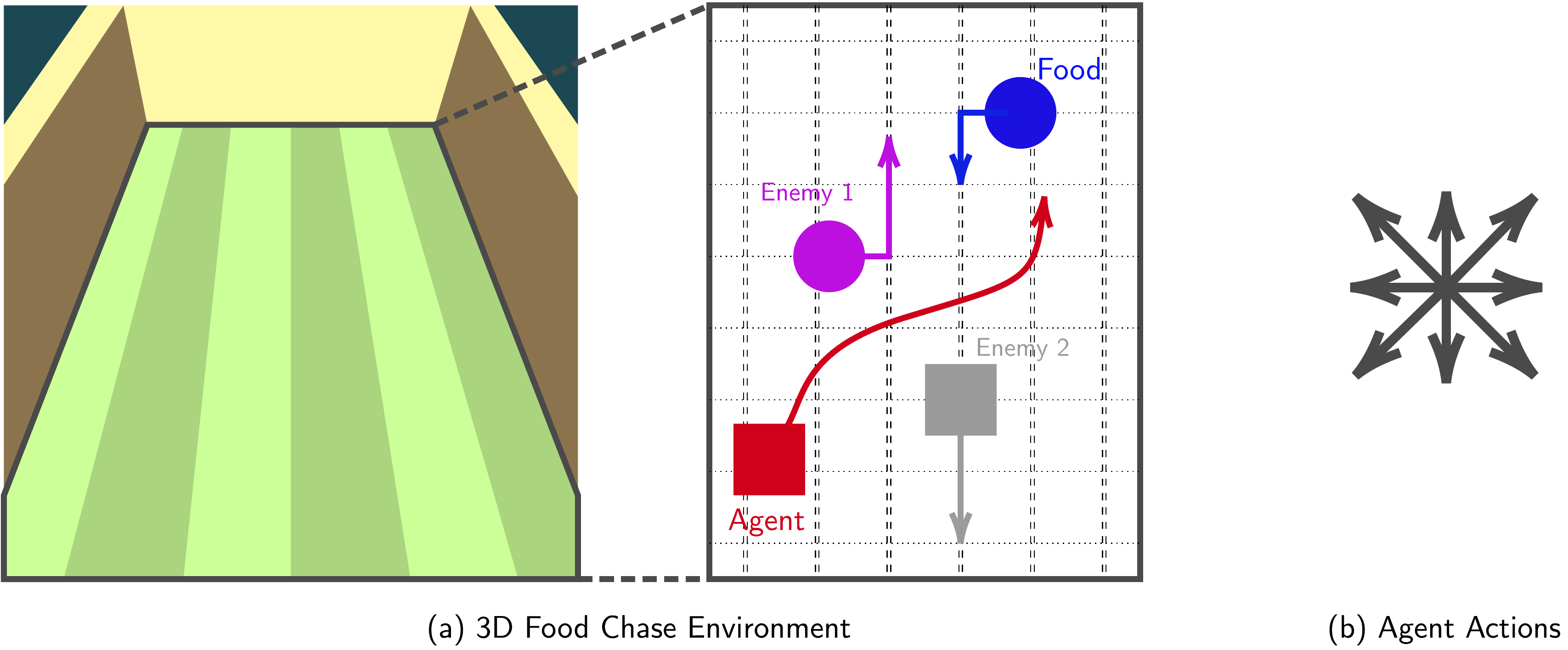}
    \caption{\textbf{3D Food Chase Game Scheme.} The game comprises of an agent, a food and two enemies. The food and the enemies move along the vertical lanes but they can randomly change to an adjacent lane at the horizontal crossings. The aim of the agent is to chase and eat the food. Once the food is eaten the food re-spawns at a random location. The actions of the agent comprise of choosing an acceleration in one of the 8 cardinal directions or choosing to not accelerate at all. There is negative reward of -2 for hitting the enemies. There is positive reward of +10 for eating the food. The reward is higher when one of the agent or the food are invisible (+20) and still higher when both of them are invisible (+30) when the food is eaten.}
    \label{fig:foodchase-game-rules}
\end{figure}

\subsection{Additional Results on the Analysis of AESMC}
\label{ax:aesmc-analysis}
\subsubsection{2D Billiards Dataset}
We further analyse AESMC for varying number of objects in the scene and multi-modality. For this we build on the 2D Billiards task as used in GSWM. For partial observability we use object flicker and for multi-modality/randomness we have two invisible vertical reflectors at $0.33$ and $0.66$ of the width of the canvas. On colliding with these reflectors, the object randomly either continues to have the same velocity or reverses its velocity in the opposite direction (reflection). 

\begin{figure}[h!]
    \centering
    \includegraphics[width=0.6\textwidth]{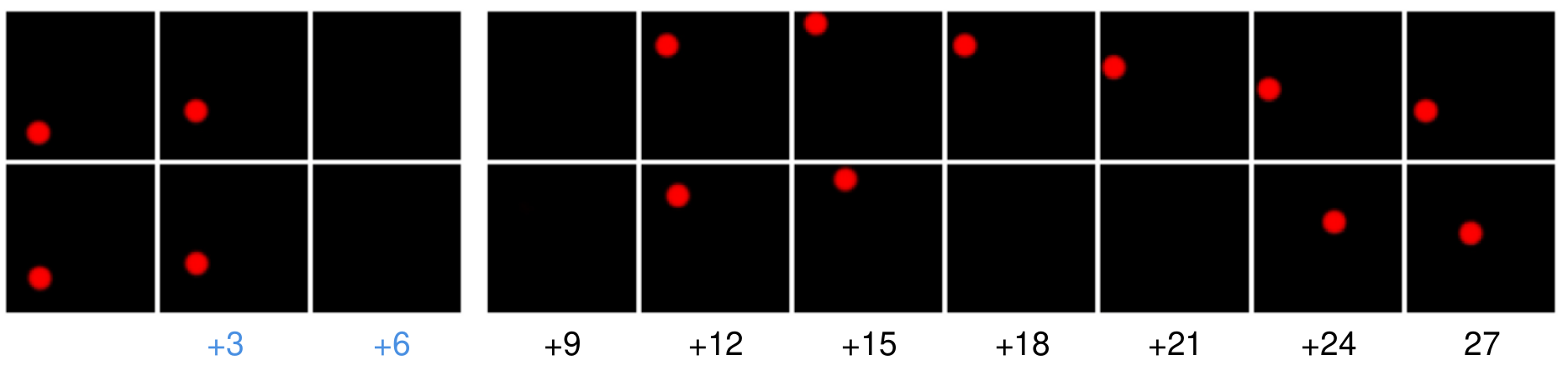}
    \caption{\textbf{AESMC Generation in 2D Billiards} We show the generation of AESMC with $N=1$ in presence of randomness (reflectors). The top row shows the ground truth and the bottom row shows the AESMC generation. Here we find that AESMC is able to learn the dynamics well and is able to perform long term rollouts.}
    \label{fig:aesmc-n1-ax}
\end{figure}

\begin{figure}[h!]
    \centering
    \includegraphics[width=0.6\textwidth]{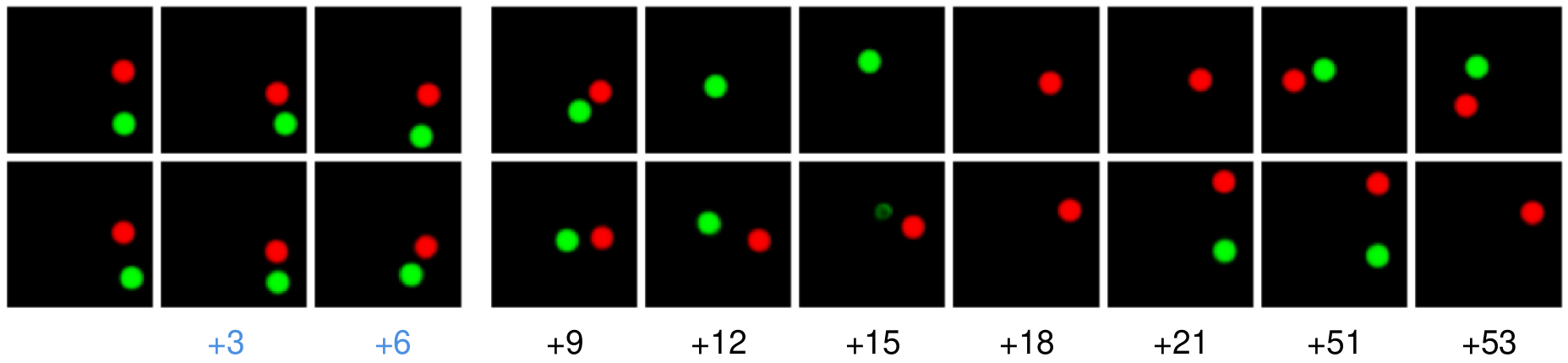}
    \caption{\textbf{AESMC Generation in 2D Billiards} We show the generation of AESMC with $N=2$ objects and in presence of randomness. The top row shows the ground truth and the bottom row shows the AESMC generation. Time steps shown in \textcolor{blue}{blue} are the conditioned images and the time steps shown in black are generation. We observe that as compared to $N=1$ case, the dynamics are less coherent. For example the motion of objects in between timesteps $+6$ and $+9$ is deterministic but there is deviation in the objects' location compared to the ground truth after time step $+9$.}
    \label{fig:aesmc-n2-ax}
\end{figure}

\begin{figure}[h!]
    \centering
    \includegraphics[width=0.55\textwidth]{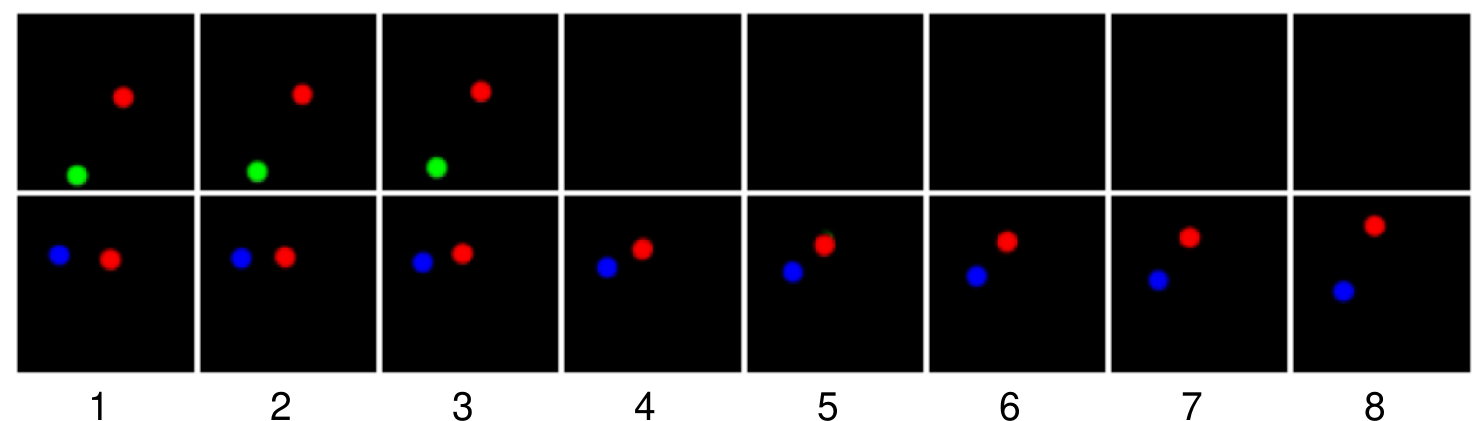}
    \caption{\textbf{AESMC Generation in 2D Billiards} We show the generation of AESMC with $N=3$ objects in presence of randomness. The top row shows the ground truth and the bottom row shows the AESMC generation. Here, we show an example in which the dynamics of red ball is not accurate. After $t=3$, the ball changes it's trajectory abruptly and moves upwards.}
    \label{fig:aesmc-n3-dynamics-ax}
\end{figure}
 
 \begin{figure}[h!]
    \centering
    \includegraphics[width=0.7\textwidth]{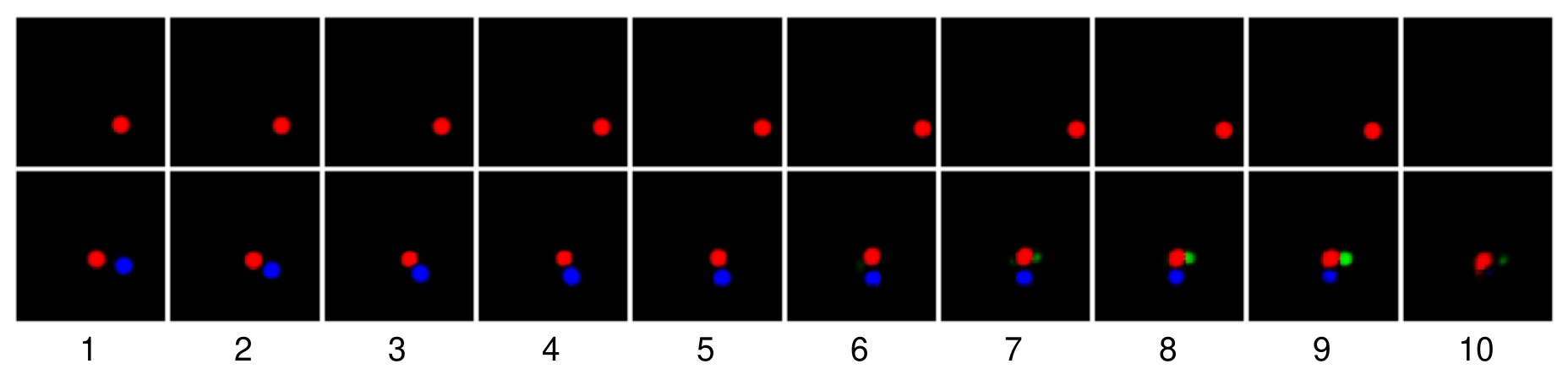}
    \caption{\textbf{AESMC Generation in 2D Billiards} We show the generation of AESMC with $N=3$ objects in presence of randomness. The top row shows the ground truth and the bottom row shows the AESMC generation. Here, we show an example in which the model generates incoherent objects at $t=8$. Additionally, the dynamics of the blue and red balls is incorrect. The red ball stops after colliding with the blue ball.}
    \label{fig:aesmc-n3-color-ax}
\end{figure}

 \begin{figure}[h!]
    \centering
     \includegraphics[width=0.7\textwidth]{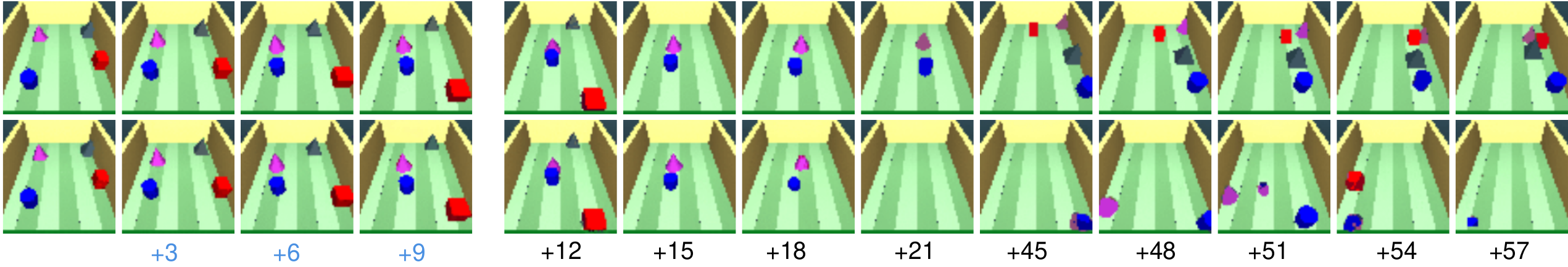}
    \caption{\textbf{AESMC Generation in 3D Task} The top row shows the ground truth and the bottom row shows the AESMC generation. We observe that for example at $t=51$, the model is confused about the position of the pink pyramid. Also, there is an abrupt of position of the blue cylinder from extreme right at $t=51$ to extreme left at $t=54$, which is incorrect.}
    \label{fig:aesmc-3d-foodchase-ax}
\end{figure}

\subsubsection{Experiment Details}
We run both AESMC and SWB with $K=10$ particles and evaluate the positional log-likelihood as described in Section \ref{sec:world_model}. In Figures \ref{fig:aesmc-n1-ax}, \ref{fig:aesmc-n2-ax}, \ref{fig:aesmc-n3-dynamics-ax}, \ref{fig:aesmc-n3-color-ax} we show qualitative results of AESMC with different objects $N=\{1, 2, 3\}$ in the presence of randomness (reflectors). We note that for smaller number of objects ($N=1$), AESMC is able to generate coherent sequences but as we increase the number of objects (with randomness present), learning of object dynamics becomes difficult and thus generating implausible frames. We note that the capacity of the image encoder and decoder in SWB and AESMC was matched and was not the bottleneck. We hypothesize that the lack of structure in case of AESMC is the reason for its incorrect dynamics. With added randomness, the complexity in modelling the dynamics increases.


\section{SPACE-based Implementation of Structured World Belief}
We implement our model by adopting the object-centric representation of SPACE and dynamics modeling of GSWM as our framework. Hence, we decompose the object states in our object files as:
\begin{align}
    \bz_{t,n}^\text{k, \text{obj}} = (\bz_{t,n}^\text{k, \text{depth}},  \bz_{t,n}^\text{k, \text{scale}},
    \bz_{t,n}^\text{k, \text{position}},
    \bz_{t,n}^\text{k, \text{velocity}},
    \bz_{t,n}^\text{k, \text{what}},
    \bz_{t,n}^\text{k, \text{dynamics}})
\end{align}
\textit{Depth} is used to resolve occlusions during rendering. \textit{Scale} is the size of the bounding box that encloses the object. \textit{Position} is the center position of the bounding box. \textit{Velocity} is the difference between the last object position and the current object position, representing the velocity. \textit{What} is a distributed vector representation that is used to decode the object glimpse that is rendered inside the bounding box. \textit{Dynamics} is a distributed vector representation which encodes noise that enables multi-modal randomness in object trajectory during generation. 

\subsection{Generation}
Here, we describe the process of doing future generation starting from a given file state $\bs_{t, 1:N}^k$. For brevity of the description of the generative process, we will drop the superscript $k$.

 \textbf{Background Generation.} We treat the background to be a special object file which is always visible. Furthermore, its position is the center of the image and its scale stretches over the entire image. We use a background module similar to GSWM \citep{gswm}. The generative process for the background takes the background object file $\bz_{t-1}^{\text{bg}}$ from the previous time-step and predicts the sufficient statistics (mean and sigma) for the new background object file. This is done as follows.
    \begin{align}
        \boldsymbol{\mu}_{t,n}^\text{bg, prior}, \boldsymbol{\sigma}_{t,n}^\text{bg, prior} &= \operatorname{MLP}_\ta^\text{bg}(\bz_{t-1}^{\text{bg}}).\nn
    \end{align}
    Then the new background state is sampled as follows.
    \begin{align}
        \bz_{t}^{k, \text{bg}} \sim \cN(\boldsymbol{\mu}_{t,n}^\text{bg, prior}, \boldsymbol{\sigma}_{t,n}^\text{bg, prior})
    \end{align}
    The background is rendered using a CNN with deconvolution layers and sigmoid output activation.
    \begin{align}
        \by_{t}^{\text{bg}} = \operatorname{CNN}^\text{RenderBG}_\ta(\bz_{t}^{\text{bg}})
    \end{align}
    We then obtain separate background contexts for each object file to condition the generation. For this, a bounding box region around the last object file position is used to crop the background RGB map.
    \begin{align}
        \by_{t,n}^{\text{bg}} = \operatorname{crop}(\by_{t}^{ \text{bg}}, \bz_{t,n}^\text{ \text{scale}} + \Delta_\text{bg, proposal}, \bz_{t,n}^\text{ \text{position}})
    \end{align}
    where $\Delta_\text{bg, proposal}$ is a hyper-parameter taken as 0.25 in this work. The cropped context is then encoded using a CNN as follows.
    \begin{align}
        \bee_{t,n}^{ \text{bg}} = \operatorname{CNN}_\ta^\text{EncodeBGProposal}(\by_{t,n}^{\text{bg}})
    \end{align}
    \textbf{Foreground Generation.} Then we perform file-file interaction of the foreground object files using a graph neural network. This approach was also taken in \cite{stove, gswm, op3}. However, these files also interact with the neighborhood region of the background through conditioning on the background context \cite{gswm} obtained in the previous step.
    \begin{align}
    \boldsymbol{e}_{t,n}^\text{file-interaction} &= \operatorname{GNN}_\ta (\bz_{\tmo,1:N}, \bh_{\tmo,1:N}, \bee_{t,1:N}^{\text{bg}}).\nn
    \end{align}
   We then compute sufficient statistics for the random variables that need to be sampled.
    \begin{align}
    \boldsymbol{\mu}_{t,n}^\text{obj}, \boldsymbol{\sigma}_{t,n}^\text{obj}, \boldsymbol{\rho}_{t,n}^\text{vis} &= \operatorname{MLP}_\ta(\boldsymbol{e}_{t,n}^\text{file-interaction}).\nn
    \end{align}
   Next, we sample the random variables.
    \begin{align}
    \bz_{t,n}^\text{vis} &\sim \text{Bernoulli}(\cdot | \boldsymbol{\rho}_{t,n}^\text{vis}),\nn\\
    \bz_{t,n}^\text{obj} &\sim \cN(\boldsymbol{\mu}_{t,n}^\text{obj}, \boldsymbol{\sigma}_{t,n}^\text{obj})
    .\nn
    \end{align}
    We update the RNN as follows.
    \begin{align}
    \bh_{t,n}^k &= \text{RNN}_\ta(\bz_{t,n}^{k}, \bh_{t-1,n}^k).\nn
    \end{align}
    We carry over the previous object ID.
    \begin{align}
    i_{t,n} &= i_{t-1,n}.\nn
    \end{align}

\subsection{Inference}

\subsubsection{Image Encoder}
\label{ax:image_encoder}

We feed the image to a CNN. We interpret the output feature cells as objects.
\begin{align}
    \boldsymbol{u}_{t, 1:G\times G}^{\text{CNN}} &= \operatorname{CNN}_\phi(\bx_t).\nn
\end{align}
where $G\times G$ is the number of output cells of the CNN encoder.
Like SPACE \citep{space}, we interpret each feature $\boldsymbol{u}_{t, m}^{\text{CNN}}$ to be composed of object presence, position and scale of object bounding box and other inferred features.
\begin{align}
    \boldsymbol{u}_{t, m}^{\text{CNN}} = (\boldsymbol{u}_{t, m}^{\text{pres},\text{CNN}}, \boldsymbol{u}_{t, m}^{\text{position},\text{CNN}},
    \boldsymbol{u}_{t, m}^{\text{scale},\text{CNN}},
    \boldsymbol{u}_{t, m}^{\text{features},\text{CNN}}).\nn
\end{align}
We pick the best $M$ features based on the value of $\boldsymbol{u}_{t, m}^{\text{pres},\text{CNN}}$. This results in the slots that we require for file-slot matching.
\begin{align}
    \boldsymbol{u}_{t, 1:M} = \operatorname{SelectTopM}(\boldsymbol{u}_{t, 1:G\times G}^{\text{CNN}}, \texttt{key}=\boldsymbol{u}_{t, 1:G\times G}^{\text{pres},\text{CNN}}, \texttt{count}=M).\nn
\end{align}
To this set of slots, we add a \texttt{null} slot when the file-slot matching need not attach to any of the detected features.
\begin{align}
    \boldsymbol{u}_{t, 0:M} = \bu_{t,0} \cup \boldsymbol{u}_{t, 1:M}.\nn
\end{align}

\subsubsection{File-Slot Attachment and Glimpse Proposal}

In this step, we want each previous object file to perform attention on one of the slots provided by the image encoder. This is done by sampling a stochastic index $m_{t,n}^k$ for each previous object file $\bs_{t,n}^k$ as described in Algorithm~\ref{algo:object_matching}.

In our SPACE-based implementation, we take the match index and obtain the proposal bounding box as:
    \begin{align}
        \bo_{t,n}^{k, \text{proposal}} = (\boldsymbol{u}_{t, m_{t,n}^{k}}^\text{position}, \boldsymbol{u}_{t, m_{t,n}^{k}}^\text{scale})
    \end{align}
    
 Using the proposal, we crop the image to get $\bx_{t,n}^{k, \text{crop}}$.
    \begin{align}
        \bx_{t,n}^{k, \text{crop}} = \operatorname{crop}(\bx_t, \bo_{t,n}^{k, \text{proposal}})
    \end{align}
We then encode this cropped patch using a CNN to get a glimpse encoding $\bee_{t,n}^{k, \text{glimpse}}$.
    \begin{align}
        \bee_{t,n}^{k, \text{glimpse}} = \operatorname{CNN}_\phi (\bx_{t,n}^{k, \text{crop}})
    \end{align}

\subsubsection{State Inference}
We then use the glimpse encoding to infer new state of the object files. This is done as follows.

\textbf{Background Inference.}
    We will first infer and render the background. The inference process takes the input image and feeds it to an encoder CNN to obtain an image encoding.
    \begin{align}
        \bee_{t}^\text{bg} = \operatorname{CNN}_\phi^\text{EncodeBG}(\bx_t)
    \end{align}
    The encoding is concatenated with the object file of the background from the previous time-step and fed to an MLP to predict the sufficient statistics (mean and sigma) for the inference distribution. 
    \begin{align}
        \boldsymbol{\mu}_{t}^{k,\text{bg, post}}, \boldsymbol{\sigma}_{t}^{k,\text{bg, post}} &= \operatorname{MLP}_\phi^\text{bg}(\bz_{t-1}^{k, \text{bg}}, \bee_{t}^\text{bg}).\nn
    \end{align}
    Then the new background state is sampled as follows.
    \begin{align}
        \bz_{t}^{k, \text{bg}} \sim \cN(\boldsymbol{\mu}_{t}^{k,\text{bg, post}}, \boldsymbol{\sigma}_{t}^{k,\text{bg, post}})
    \end{align}
    Lastly, we render the background using a CNN with deconvolution layers and sigmoid output activation.
    \begin{align}
        \by_{t}^{\text{bg}} = \operatorname{CNN}^\text{RenderBG}_\ta(\bz_{t}^{\text{bg}})
    \end{align}

    We then obtain separate background contexts for each object file to condition the inference. For this, a bounding box region around the last object file position is used to crop the background RGB map.
    \begin{align}
        \by_{t,n}^{k, \text{bg}} = \operatorname{crop}(\by_{t}^{ \text{bg}}, \bz_{t,n}^\text{k, \text{scale}} + \Delta_\text{bg, proposal}, \bz_{t,n}^\text{k, \text{position}})
    \end{align}
    where $\Delta_\text{bg, proposal}$ is a hyper-parameter taken as 0.25 in this work. The cropped context is then encoded using a CNN as follows.
    \begin{align}
        \bee_{t,n}^{k, \text{bg}} = \operatorname{CNN}_\ta^\text{EncodeBGProposal}(\by_{t,n}^{k, \text{bg}})
    \end{align}
    \textbf{Foreground Inference:} We first perform file-file interaction for the foreground object files using a graph neural network. This approach was also taken in \cite{stove, gswm, op3} to model object interactions such as collisions in physical dynamics. This interaction also takes into account the situation of the objects in the context of the background.
    \begin{align}
    \boldsymbol{e}_{t,n}^{k,\text{file-interaction}} &= \operatorname{GNN}_\ta (\bz_{\tmo,1:N}^k, \bh_{\tmo,1:N}^k, \bee_{t,1:N}^{k, \text{bg}}).\nn
    \end{align}
     We then compute sufficient statistics for the random variables that need to be sampled.
    \begin{align}
    \boldsymbol{\mu}_{t,n}^{k,\text{obj, post}}, \boldsymbol{\sigma}_{t,n}^{k,\text{obj, post}}, \boldsymbol{\rho}_{t,n}^{k,\text{vis, post}} &= \operatorname{MLP}_\phi(\boldsymbol{e}_{t,n}^{k,\text{file-interaction}}, \bee_{t,n}^{k, \text{glimpse}}).\nn
    \end{align}
    We then compute sufficient statistics for the prior distribution.
    \begin{align}
    \boldsymbol{\mu}_{t,n}^{k,\text{obj, prior}}, \boldsymbol{\sigma}_{t,n}^{k,\text{obj, prior}}, \boldsymbol{\rho}_{t,n}^{k,\text{vis, prior}} &= \operatorname{MLP}_\ta(\boldsymbol{e}_{t,n}^{k,\text{file-interaction}}).\nn
    \end{align} 
    Next, we sample the random variables.
    \begin{align}
    \bz_{t,n}^{k,\text{vis}} &\sim \text{Bernoulli}(\cdot | \boldsymbol{\rho}_{t,n}^{k,\text{vis, post}}).\nn
    \end{align}
    Next, we sample the remaining random variables. We perform object-wise imagination or bottom-up inference depending on whether the object is visible or not.
    \begin{align}
    \bz_{t,n}^{k,\text{obj}} &\sim \cN(\cdot | \boldsymbol{\mu}_{t,n}^{k,\text{obj, post}}, \boldsymbol{\sigma}_{t,n}^{k,\text{obj, post}})^{\bz_{t,n}^{k,\text{vis}}} \cN(\cdot | {\boldsymbol{\mu}}_{t,n}^{k,\text{obj, prior}}, {\boldsymbol{\sigma}_{t,n}}^{k,\text{obj, prior}})^{1 - \bz_{t,n}^{k,\text{vis}}}.\nn
    \end{align}
     We update the RNN as follows using the sampled object states.
    \begin{align}
    \bh_{t,n}^k &= \text{RNN}_\ta(\bz_{t,n}^{k}, \bh_{t-1,n}^k).\nn
    \end{align}
    To update the ID, if the file is set to visible for the first time, we assign a new ID to the file. Otherwise, we simply carry over the previous ID. That is inactive file remains inactive. And active file remains active with the same ID.
    \begin{align}
    i_{t,n}^k &= \begin{cases}
    \texttt{new\_id()} & \text{if } \bz_{t,n}^{k, \text{vis}} = 1 \text{ and } i_{t-1,n}^k = \texttt{null} \nn\\
    i_{t-1,n}^k & \text{otherwise}\nn
    \end{cases}
    \end{align}

\subsubsection{Position Prediction via Velocity Offset}
Similarly to STOVE \citep{stove} and GSWM \citep{gswm}, we predict the position (both during inference and generation) by adding \textit{velocity} $\bz_{t,n}^{k, \text{velocity}}$ to the previous object position. All other attributes are directly predicted without adding offsets.
\begin{align}
    \bz_{t,n}^{k, \text{position}} =  \bz_{t-1,n}^{k, \text{position}} +  \bz_{t,n}^{k, \text{velocity}}
\end{align}

\subsubsection{Rendering}
Our final image is rendered similarly to SPACE and GSWM \citep{space, gswm}. Rendering process takes the object files for the background $\bz_t^{k, \text{bg}}$ and the foreground objects $\bz_{t, 1:N}^{k}$ and returns pixel-wise means of the RGB values in the final image. The sigma is set as a hyper-parameter.
\begin{align}
    \boldsymbol{\mu}_t^{k,\text{rendered}} &= \operatorname{render}_\ta(\bz_t^{k, \text{bg}}, \bz_{t, 1:N}^{k})\\
    \boldsymbol{\sigma}_t^{k,\text{rendered}} &= \text{Set as hyperparameter.}
\end{align}

\subsection{Training}
In this section, we describe details related to training that were not described in the main text.
\subsubsection{Curriculum}
For training iterations up to 3000, we train the image encoder using the auto-encoding objective of SPACE. This allows image encoder to learn to detect objects in images. After 3000 iterations, the SPACE objective is removed. Hereafter, the output of image encoder is directly treated as $M$ slots which are provided as input for file-slot matching as described in Appendix~\ref{ax:image_encoder} and in the main text. The encoder is jointly trained with the rest of the model.


\subsubsection{Soft-Resampling}
 To prevent low-weight particles from propagating, we resample the particles based on the particle weights. However, when we resample the particles from the distribution $q^\text{resampler}(k) = {w}_t^k$, we obtain a non-differentiable sampling process which prevents gradient from flowing back. This problem is commonly addressed by applying soft-resampling \citep{soft-resampling} as follows:
\begin{align}
    q^\text{resampler}(k) &= \alpha {w}_t^k + (1 - \alpha) \frac{1}{K}\\
    \quad \text{pa}(k) &\sim q^\text{resampler}(\cdot), \forall k\\
    \bs_{t,n}^k \leftarrow {\bs}_{t,n}^{\text{pa}(k)} &\text{ and }
    w_t^k \leftarrow \frac{{w}_t^{\text{pa}(k)}}{\alpha {w}_t^{\text{pa}(k)} + (1 - \alpha)\frac{1}{K}}
\end{align}
where $\alpha$ is a trade-off parameter. We choose $\alpha = 0.6$ during training.


\newpage
\subsubsection{Hyper-Parameters}
\begin{table}[h!]
\centering
\begin{tabular}{@{}llll@{}}
\toprule
Parameter                     & 2D Branching Sprites     & 2D Maze                        & 3D Food Chase        \\ \midrule
Image Width/Height & 64 & 64 & 64 \\
$N$ & 4 & 8 & 8 \\
$K$ & 10 & 10 & 10 \\
Learning Rate & 0.0001 & 0.0001 & 0.0001 \\
Optimizer & Adam & Adam & Adam \\
\midrule
Image Encoder Grid Cells & $8\times 8$ & $16 \times 16$ & $16 \times 16$ \\
Background Module & Off & On & On \\
Glimpse Width/Height & 16 & 16 & 16 \\
Glimpse Encoding Size & 128 & 128 & 128 \\
\midrule
RNN hidden state size & 128 & 128 & 128 \\
\texttt{depth} attribute size & 1 & 1 & 1 \\
\texttt{scale} attribute size & 2 & 2 & 2 \\
\texttt{position} attribute size & 2 & 2 & 2 \\
\texttt{offset} attribute size & 2 & 2 & 2 \\
\texttt{what} attribute size & 32 & 32 & 32 \\
\texttt{dynamics} attribute size & 8 & 8 & 8 \\
\midrule
Reconstruction Sigma & 0.05 & 0.075 & 0.1 \\
\bottomrule
\end{tabular}
\caption{Model Specifications}
\end{table}

\end{appendices}
\end{document}